\documentclass[3p,10pt,times,authoryear]{elsarticle}

\usepackage{amssymb}

\usepackage{amsthm}
\usepackage{amsmath} 
\usepackage[caption=false]{subfig}
\usepackage{setspace}
\usepackage{graphicx}
\usepackage{float}
\usepackage{upgreek}
\usepackage{amsmath} 
\usepackage{gensymb} 
\graphicspath{ {fig/} }

\usepackage[font=small,labelfont={bf},tableposition=top]{caption}
\usepackage{blindtext}
\journal{ISTVS 15$^{th}$ European-African Regional Conference, Prague, Czech Republic, September 9--11, 2019}

\begin{document}

\begin{frontmatter}



\title{THE EFFECTS OF INCREASING VELOCITY ON THE TRACTIVE PERFORMANCE OF PLANETARY ROVERS}


\author[1,2]{David Rodr\'iguez-Mart\'inez\corref{cor1}}
\ead{davidrm@dc.tohoku.ac.jp}
\cortext[cor1]{Corresponding author.}

\author[3]{Fabian Buse}
\author[2]{Michel Van Winnendael}
\author[1]{Kazuya Yoshida}

\address[1]{Department of Aerospace Engineering,
Tohoku University, 6-6-01, Aramaki Aoba, Sendai, 980-8570, Japan}
\address[2]{European Space Agency, ESTEC, Keplerlaan 1, 2200AG, Noordwijk, The Netherlands}
\address[3]{Institute of System Dynamics and Control, German Aerospace Center (DLR), M{\"u}nchener Str. 20, 82234 We{\ss}ling, Germany}

\tnotetext[label1]{The authors are solely responsible for the content of this technical presentation. The technical presentation does not necessarily reflect the official position of the International Society for Terrain Vehicle Systems (ISTVS), and its printing and distribution does not constitute an endorsement of views which may be expressed. Technical presentations are not subject to the formal peer review process by ISTVS editorial committees; therefore, they are not to be presented as refereed publications. Citation of this work should state that it is from an ISTVS meeting paper. EXAMPLE: Author's Last Name, Initials. 2019. Title of Presentation. The ISTVS 15th European-African Regional Conference, Prague, Czech Republic. For information about securing permission to reprint or reproduce a technical presentation, please contact ISTVS at 603-646-4405 (72 Lyme Road, Hanover, NH 03755-1290 USA)}

\begin{abstract}
An emerging paradigm is being embraced in the conceptualization of future planetary exploration missions. Ambitious objectives and increasingly demanding mission constraints stress the importance associated with faster surface mobility. Driving speeds approaching or surpassing 1\,m/s have been rarely used and their effect on performance is today unclear. This study presents experimental evidence and preliminary observations on the impact that increasing velocity has on the tractive performance of planetary rovers. Single-wheel driving tests were conducted using two different metallic, grousered wheels ---one rigid and one flexible--- over two different soils, olivine sand and $\textrm{CaCO}_3$-based silty soil. Experiments were conducted at speeds between 0.01--1\,m/s throughout an ample range of slip ratios (5--90\%). Three performance metrics were evaluated: drawbar pull coefficient, wheel sinkage, and tractive efficiency. Results showed similar data trends among all the cases investigated. Drawbar pull and tractive efficiency considerably decreased for speeds beyond 0.2\,m/s. Wheel sinkage, unlike what published evidence suggested, increased with increasing velocities. The flexible wheel performed the best at 1\,m/s, exhibiting 2 times higher drawbar pull and efficiency with 18\% lower sinkage under low slip conditions. Although similar data trends were obtained, a different wheel-soil interactive behavior was observed when driving over the different soils. Overall, despite the performance reduction experienced at higher velocities, a speed in the range of 0.2--0.3\,m/s would enable 5--10 times faster traverses, compared to current rovers driving capability, while only diminishing drawbar pull and efficiency by 7\%. The measurements collected and the analysis presented here lay the groundwork for initial stages in the development of new locomotion subsystems for planetary surface exploration. At the same time, these data support the creation of velocity-dependent traction models required in the later stages of the development and subsequent operation of future, fast-moving planetary rovers. 
\end{abstract}


\end{frontmatter}


\section{Introduction}
\label{sec:intro}
\noindent
Space exploration has been driven, to a great extent, by the use and advance of robotic technology. Planetary mobile robotic systems, popularly known as rovers, have been used in eight different missions to the Moon and Mars. They have explored, as of today, a total of approximately 214\,km, and unveiled critical information shaping our current understanding of the history and evolution of the Solar System. The level of experience and sophistication acquired over the years in the design and operation of mobile robots makes them the preferred tool of use under the menacing environmental conditions of space. 

Rovers are designed with the objective to maximize reliability while minimizing mass and mechanical complexity. A 6-wheeled chassis configuration with the well-known rocker-bogie suspension ---a passive-rolling, self-balancing suspension system--- and metallic grousered wheels comprised the go-to solution in most of the aforementioned expeditions. The robustness and reliability of this solution to effectively cope with the variety of terrains often object of extraterrestrial exploration have been extensively proven \citep{Malenkov2015}. However, its application has been limited to driving speeds that rarely surpass a few cm/s. Reasons for such low velocities are multifold: from safety concerns motivated by extreme environmental conditions, reliability, and low situational awareness; to limitations driven by technical constraints, such as available computational power and bandwidth \citep{Seeni2010}.

Today's space scene, in which exploration is likely to become a shared endeavor between government agencies and private organizations, may call for a redefinition of the approaches once undertaken. Large surface coverage in the order of hundreds of kilometers per mission \citep{Steenstra2016}, limited time windows for exploration \citep{Potts2015}, and increasingly more restricted budgets and demanding mission constraints make evident the importance associated with surface mobility. Present approaches to extraterrestrial surface exploration require the majority of the time spent on the ground to be allotted to the migration between the different regions of interest. Additionally, current limitations on the computational capabilities of radiation-hardened processing units lead to low driving duty-cycles ---i.e., the amount of time spent in motion relative to the total transit duration. A faster mobility would drastically increase the likelihood of these new missions ever being pursued, allowing the majority of the mission time to be instead invested in value-driven operations; subsequently increasing the return on investment of a given mission.

Faster surface mobility can be accomplished by either augmenting driving duty-cycles through the improvement of communication, navigation, and sensing technologies, or by increasing absolute speed values, in which our work focuses. This text represents the first step in the development of a rover locomotion subsystem capable of traveling at high speed. In this context, high-speed is used to refer to driving speeds that approach or surpass 1\,m/s; i.e, ten to nearly a hundred times faster than currently conceived rover operating velocities. 

In extraterrestrial exploration, multiple challenges arise when increased velocities are considered. From the experience acquired in the early days of lunar exploration, we now know that the combination of faster driving velocities, a reduced-gravity field, and an irregular terrain may be source of dangerous situations. The Russian \emph{lunokhods}, used during the Luna 17 and Luna 21 missions, were capable of moving at speeds up to 0.5\,m/s \citep{Malenkov2016a}. NASA's Lunar Roving Vehicle, a two-crew piloted rover used during the last three Apollo Moon missions, was reported to have reached a top-speed of 5\,m/s \citep{NASA1973a}: the fastest any vehicle has been ever driven on the lunar surface. In both cases, telemetry was unrecorded or inaccessible. Post-mission reports declared the constant bouncing of the suspension as one of the major causes for the reduced maneuverability experienced when driving at high speeds. A high risk of entrapment in the loosely compacted lunar regolith was also reported. Poor illumination conditions tended to impair visibility by making difficult to discriminate between potential obstacles and changes in the terrain relief; while lunar dust mitigation measures were reported critical for the correct performance of the locomotion subsystem.

In the realm of ground applications, several authors have analyzed the performance of airplanes, military vehicles, and farming trucks operating at high speeds under off-road conditions. Despite the circumstances associated with these applications ---i.e., top speed, wheel load, and wheel and soil types--- not being particularly suited to the constraints of planetary exploration, several observations are worth noticing. In the operation of aircraft over unprepared runways, both rut depth and rolling resistance was observed to decrease with increasing speeds at the lowest and highest velocity ranges; i.e., from 0 to about 10\,m/s and over 18\,m/s, respectively (these numbers were shown to largely vary based on tire pressure and the type of soil considered)\citep{Bock1960,Crenshaw1972}. At the highest speed range, the improvement of performance was associated with a phenomenon called \emph{soil lift}. Analogous to hydrodynamic lift effects, soil inertia would increase with increasing wheel velocities up to the point where the inertial lift produced by the soil is high enough to overcome drag forces.  The lowest velocity at which this phenomenon was apparently triggered was referred to as \emph{planing velocity} \citep{Crenshaw1972}. In testing military vehicles, similar effects where reported \citep{Coutermarsh2007}. The role of soil inertia and penetration rates have been also studied in the field of soil mechanics \citep{Pope1969,Grahn1991} and briefly applied to the operations of rigid wheels over sandy and clay soils \citep{Pope1971}. Here similar trends were observed. In the former, equal pressure levels were measured at lower sinkages for higher penetration rates. In the later, lower sinkages were measured for increasing velocities. In the field of agriculture, benefits in the treatment of farm lands were also reported in the form of lower compaction levels of the soil when trucks were operated at higher speeds \citep{Carman1994, Taghavifar2013, Taghavifar2014}.

These experiences, along with foreseeable trends in the field of extraterrestrial surface exploration, justify the need to thoroughly characterize the performance of rovers operating at increasingly higher speeds under conditions that better resemble that of the surface of the Moon and Mars. To the best of our knowledge, this paper presents the first experimental evidence of the effects that increasing velocity has on the tractive performance of planetary rovers. By means of single-wheel characterization tests, we aim to aid in the comprehensive understanding of the potential of fast driving speeds while answering the following questions: 

\begin{enumerate}
\item{How are drawbar pull, wheel sinkage, and tractive efficiency ultimately affected by increasing speeds at varying slip ratios?}
\item{Is soil lift triggered within the range of velocities and operating conditions evaluated?}
\item{What are the distinct behaviors observed when operating at high speed over different types of soils?}
\item{How do the results obtained relate to what was previously observed in the subject of fast off-road mobility?}
\item{What are the challenges and limitations of conducting wheel-soil characterization tests at higher speeds?}
\item{Given the conditions of the experiments conducted, what speed range would be considered in order to maximize performance?}
\end{enumerate}

\section{Materials and Methods}
\label{sec:method}
\noindent
A velocity-dependent characterization of the wheel-soil interaction was conducted. Due to the higher speeds contemplated in our experiments, conventional single wheel testbeds (SWTBs) could not be used. Longer soil bins, more powerful drive units, and sensors capable of storing data at higher rates were required. For this, we approached the Institute of System Dynamics and Control of the German Aerospace Center (DLR), where an automated facility for the terramechanical characterization of robots had been recently developed. 

\subsection{The Terramechanics Robotics Locomotion Lab}
\label{subsec:troll}
\begin{figure}[!hbt]
\centering
\includegraphics[width=0.8\linewidth]{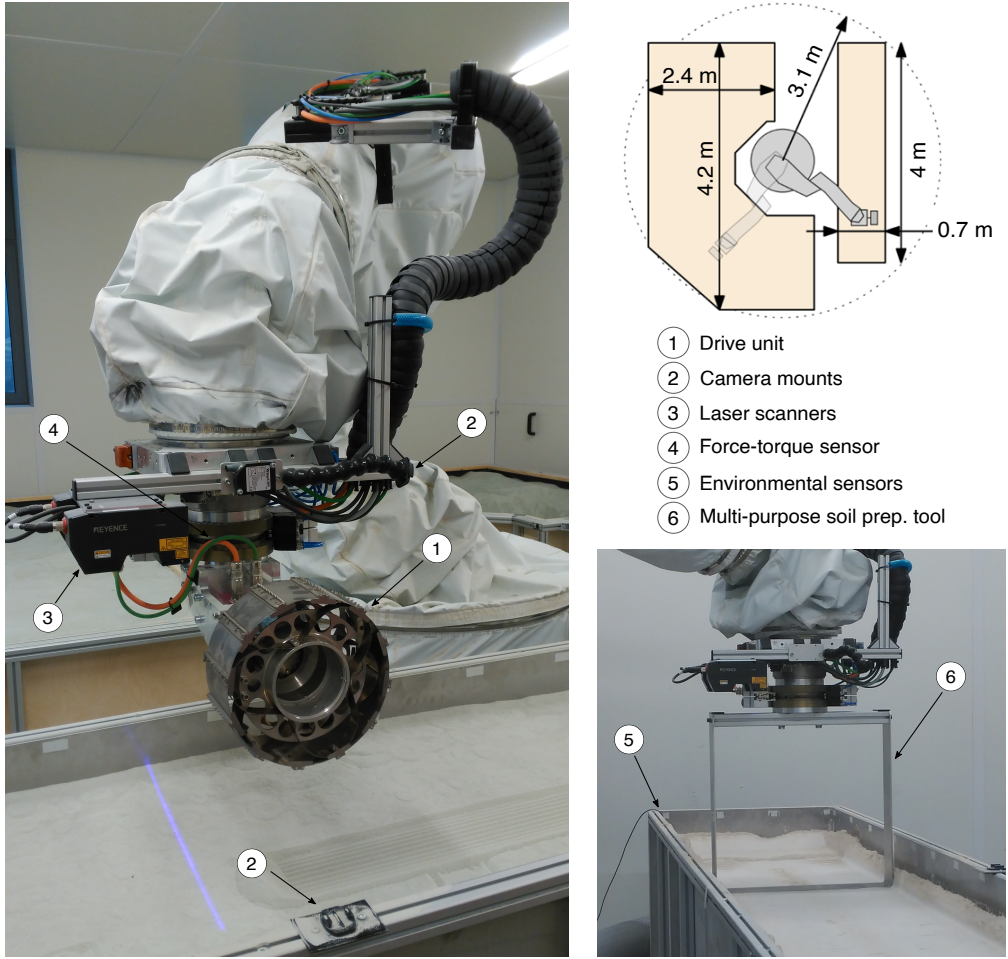}  
\caption{The Terramechanics Robotics Locomotion Lab (TROLL).}
\label{fig:troll}
\end{figure}

\noindent
The so-called Terramechanics Robotics Locomotion Lab (TROLL) makes use of a KUKA KR210 R3100 Ultra robotic arm to provide and extend the functionality of conventional SWTBs. The facility includes a closed, dust-protected test area where the robot arm stands surrounded by two large soil bins (Fig.~\ref{fig:troll}). These are designed to host a range of multiple operations, benefiting from the wide range of motion of the robot arm. The 4-meter-long straight soil bin was the only one used during the experiments presented here. The enclosed facility is also equipped with an array of sensors to measure temperature, relative air and soil humidity levels, which aid to maintain controlled ambient conditions. Directly attached to the robot flange is a sensor mount, which consists of a ATI Omega160 IP65/IP68 SI-2500-400 force--torque sensor, a set of two Keyence LJ-7300 laser scanners, and camera holders to record the experiments. If required, a complete 3D reconstruction of the surface geometry can be obtained by combining the resulting information from the laser scanners with the recorded position of the robot arm. Attached to the sensor mount is a tool changer to which different end effectors can be mounted: namely, a multi-purpose soil preparation tool and a drive unit (Fig.~\ref{fig:troll}). Different operations can be automatically performed by the soil preparation tool based on its relative position to the ground. The specific preparation methods used in the experiments are described in Sect.~\ref{subsubsec:soilPrep}. The drive unit, on the other hand, comprises a Harmonic Drive CHA-14A-100 and attachment points where wheels of diameters between 200 to 400\,mm can be mounted. A more detailed description of the design, operation, and control of the TROLL facility can be found in \cite{Buse2018}. Additionally, a Sony NEX-FS700 mounting a 18-200 mm f/3.5--6.3 power zoom lens was used to record 4K video at 480\,fps of the the wheel operation during the experiments.

\subsection{Tests Description}
\label{subsec:expDescription}
\noindent
The test campaign comprised multiple case studies. Each case study was selected to favor a comparative analysis and to maximize the range of conditions under which high-speed driving was tested. Results from three of these cases are presented in this text, details of which are provided in Table~\ref{table:cases}. 

\begin{table}[!hbt]
\caption{Details of case studies 1--3}
\label{table:cases}
\begin{tabular}{p{1.7cm}p{1.8cm}p{2.3cm}p{2cm}p{2cm}p{2.3cm}p{1.5cm}} 
\hline
Case study & Wheel Type & Diameter [mm] & Width [mm] & Grouser Type & No. of Grousers & Soil Type \\
\hline
1 & \textbf{Rigid} & 250 & 112 & Toothed & 12 & RMCS14 \\ 
2 & \textbf{Flexible} & 250 & 112 & Toothed & 12 & RMCS14 \\ 
3 & Rigid & 250 & 112 & Toothed & 12 & \textbf{RMCS13} \\  
\hline
\end{tabular}
\end{table}

\begin{figure}[!hbt]
\includegraphics[width=\linewidth]{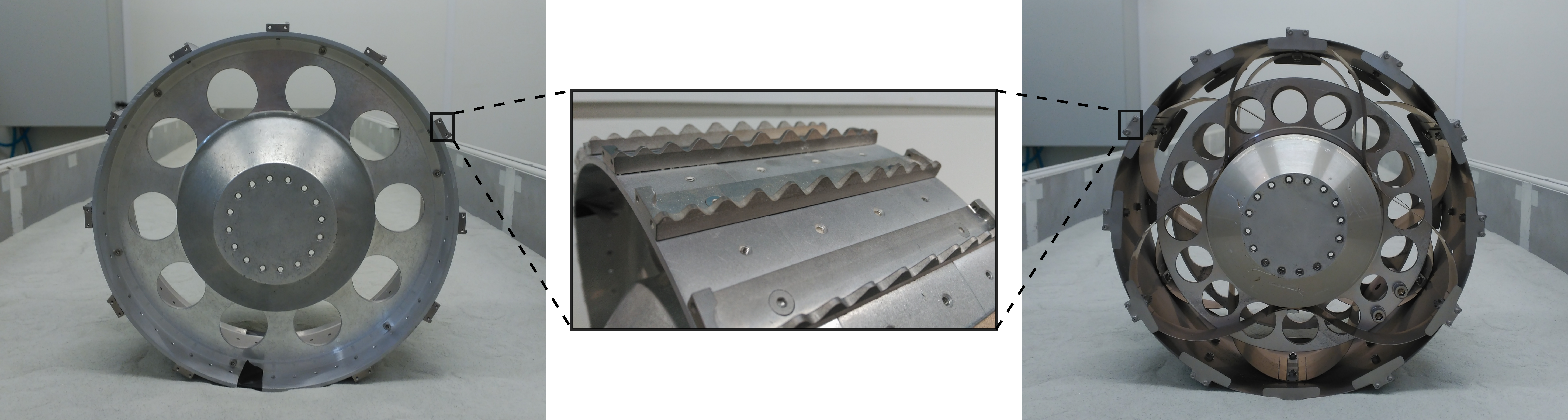}
\caption{The two wheels used in case studies 1 (left wheel) and 2 (right wheel) and a detail depiction of the toothed grousers.}
\label{fig:wheels}
\end{figure}

Experiments were initially conducted on a rigid metallic wheel of 250\,mm in diameter, 112\,mm wide, and 12 toothed grousers. This first case study was selected as the baseline to which the rest of the cases could be compared. Case study 2 featured a flexible metallic wheel of the same size, grouser type, and number of grousers. This same wheel design and grouser configuration will be used by the recently named Rosalind Franklin rover during the European Space Agency's ExoMars Mission expected to be launched in 2020. Pictures of both wheels and a detailed depiction of the design of the grousers are displayed in Fig.~\ref{fig:wheels}.

The RMCS14 soil simulant was used in these two first case studies. RMCS14 is made out of olivine sand grains selected to provide a grain size distribution that resembles that of quartz-based Martian soils if only without the silty and clay particles. In the third case, the velocity-dependent performance of our baseline wheel was evaluated under a different type of soil: the RMCS13 simulant, a calcium carbonate based material ($\textrm{CaCO}_3$-based). Due to its slight cohesiveness and fine grain size distribution, this soil simulant resembles the worst behavioral response expected of soft lunar regolith \citep{Buse2018}. Specific details of each simulant can be found in Table~\ref{table:soils} along with the ambient conditions these soils were exposed to during the experiments. A wheel load of 100\,N, which can be thought of as representative of a medium-class 6-wheeled rover (mass $\sim$ 350\,kg) under lunar gravity, was applied in all the cases under study. 

\begin{table}[!hbt]
\caption{Soil simulants characteristics and environmental conditions}
\label{table:soils}
\begin{tabular}{p{5.7cm}p{5.7cm}} 
\hline
RMCS13 & RMCS14 \\
\hline
$\textrm{CaCO}_3$-based & Olivine (quartz subsitute, no free $\textrm{SiO}_2$) \\ 
1--30\,$\upmu$m, $\sim$ 38º $^a$ & 90--700\,$\upmu$m, $\sim$ 35º $^a$\\ 
Worst case of soft lunar regolith  & Comparable to fine-grained martian soils \\ 
26.5ºC, 1.9\% $^b$ & 24.5ºC, 3.3\% $^b$\\
\hline
\end{tabular}
  \\
  $^a$ Grain size and angle of internal friction. $^b$ Average soil temperature and humidity.
\end{table}

\subsection{Experimental Process}
\label{subsec:expProcedure}
\noindent
The procedure followed on each experiment consisted of five different phases: soil preparation, tool exchange, initial positioning, driving, and surface scan. The following sections briefly describe each of these phases. A detailed description of the experimental process can be found in \cite{Buse2019}.

\subsubsection{Soil Preparation}
\label{subsubsec:soilPrep}
\noindent
The manual labor required in the soil preparation methods frequently used with conventional SWTBs makes them time consuming and prone to inconsistencies. Taking advantage of the level of automation of the facility, specific routines for the preparation of each soil were developed. Four different operations were pre-programmed into the TROLL controller: fluidization, sand mixing, leveling, and compression. The fluidization method takes advantage of the tendency of certain soils to present a liquid-like behavior when subjected to increased strain rates. By means of a double rack of PVC tubes installed at the bottom of the soil bin, pressurized air is injected to loosen and mix the soil, removing, in a matter of seconds, the track left by the wheel and any crevices on the surface.  Sand mixing, leveling, and compression make use of the multi-purpose tool previously described. In the case of sand mixing, the tool is inserted deep into the soil (about 20\,cm) and moved back and forth along the the soil bin. Just like plowing, sand mix helps to turn over superficial layers of soil, removing any traces left by the wheel and braking up top layers of compressed soil. The leveling program places the multi-purpose tool at a given angle relative to the surface and by using the frame as a blade, the tool is dragged along the surface. Lastly, compression constantly presses the tool flat onto the surface along the full length of the soil bin while exerting a pre-set pressure down to a given depth. 

For the two types of soils used in the experiments, a different combination of these operations was required. In the case of the RMCS14, fluidization followed by two leveling passes was found to be the optimal sequence. It provided the best balance between consistent measurements and reduced preparation times. The use of fluidization on its own, although fast, created an uneven and cratered surface due to the constant release of air and bubbling of the sand. Two leveling passes helped in creating a perfectly even surface and removing any potential air trapped in the superficial layers of the soil. Using this sequence of operations, approximately 3 minutes was required for the RMCS14 to be fully prepared for a new experiment. It is worth highlighting that although compression of the surface slightly improved stability of the wheel trajectory over RMCS14, the extra time required on each experiment did not justify its use in this case. On the other hand, RMCS13 required a much more elaborate preparation process. Given the slight cohesiveness and fine grain size distribution of this soil, fluidization tended to form cylindrical air channels \citep{Buse2019}, releasing the air directly outside of the bin and precluding its use. Instead, the soil required to be first mixed, followed by at least four leveling passes at decreasing depths and ended with two compression passes to consolidate the surface enough to sustain the loaded wheel. This sequence of operations took almost three times longer than that of the RMCS14. 


\subsubsection{Tool Exchange and Initial Positioning}
\label{subsubsec:toolChange}
\noindent
After the soil was adequately conditioned, the robot arm was commanded to exchange the multi-purpose soil preparation tool by the drive unit. This was then transferred to a point above the initial surface position for the experiment, where the wheel was turned to guarantee the same grouser orientation on every experiment and the different sensors were initiated and calibrated. Only vertical motion of the robot flange was enable as the wheel was being placed on the ground and loaded by the force controller with the selected wheel load. The trajectories followed by the robot arm were optimized to speed up the process between experiments. 

\subsubsection{Forced-slip Test}
\label{subsubsec:test}
\noindent
Once the desired wheel load was reached, the experiment was run given the preset input parameters: speed, wheel load, slip ratio, and driving distance. For every case study, experiments were conducted at 0.01, 0.1, 0.2, 0.4, 0.6, 0.8, and 1\,m/s. Each speed tested was measured under two different slip conditions representative of low and high slip conditions: namely, 20\% and 60\%. In some cases, experimental evidence from a wider range of slip ratios was collected to better characterize the wheel-soil interaction and properly define inflection and turning points in the data. An initial acceleration period of 0.75\,s was used. The distance selected for the experiments was 3\,m to avoid potential disturbances caused by the front and back walls of the soil bin. 

\subsubsection{Scan of the Surface}
\label{subsubsec:scan}
\noindent
At the end of the experiment, the wheel was lifted and an automated surface scan was conducted. The data recorded allow us to reconstruct the complete trail left by the wheel along the surface if needed.

\subsubsection{Monitoring Soil Conditions}
\label{subsubsec:monitoring}
\begin{figure}[!hbt]
\centering
\includegraphics[width=0.4\linewidth]{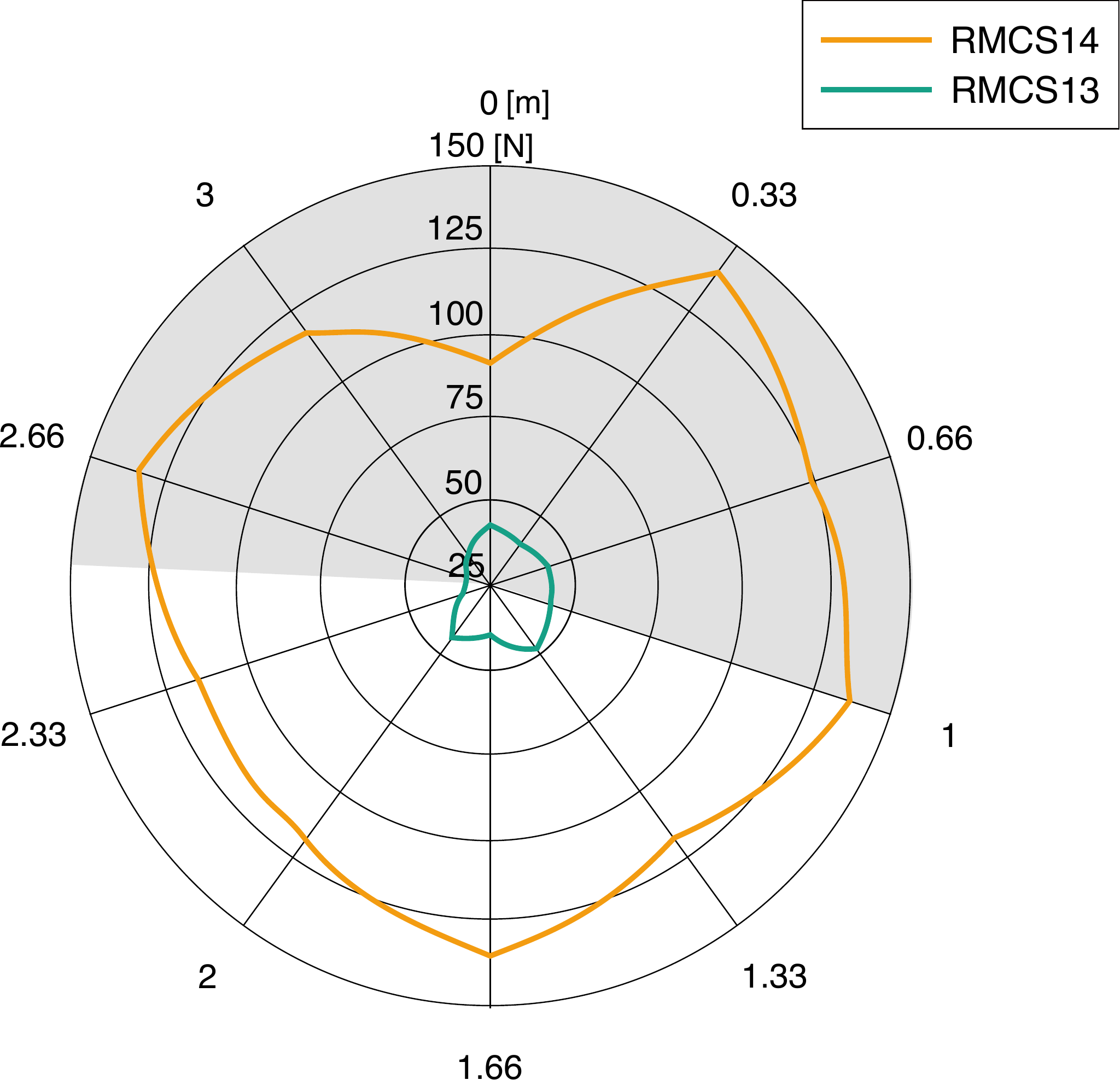}  
\caption{Penetrometer tests results along the length of the soil bin for both, RMCS14 and RMCS13 soil simulants. Penetrations were performed at 0.02\,m/s up to a depth of 0.2\,m. Radial axes define the distance of independent testing spots and the vertical axis describes the magnitude of the contact force. The gray-shaded area represents transient-state regions.}
\label{fig:charact}
\end{figure}

\noindent
Given the length of the soil bin, maintaining the soil in a homogeneous state at any given point was crucial and often a challenge. Penetrometer tests were frequently conducted in ten different spots equally distant from each other to verify that a consistent soil state was achieved. Figure~\ref{fig:charact} displays the variation of measurements for the maximum normal load recorded at a depth of 0.2\,m and a constant penetration rate of 0.02\,m/s for both soils used. The gray-shaded areas reflect sections of the soil bin from which data were discarded (refer to Sect.~\ref{subsec:dataProcessing}). In addition, multiple runs of the same experimental configuration ---i.e, same case study, velocity, and slip ratio-- were conducted to rapidly verify the state of the terrain and validate the repeatability of the experiments.

\subsection{Measurements and Data Collection}
\label{subsec:dataCollection}
\noindent
The exchange of tools, the different routines for soil preparation, and the required experimental conditions were programmed and fully automated. The KUKA Roboter Sensor Interface (RSI) allows full control of the robot arm and provides the telemetry required to track the position of the arm at all times. This, together with other selected parameters of the operation of the TROLL facility, could be constantly monitored through visual displays. A constant wheel load and a forced slip ratio can be imposed by the arm thanks to the 4\,ms control cycle of the RSI. As will be shown later, this imposed certain limitations when speeds approached 1\,m/s or very high slip ratios were tested (over 80\%). Experimental results, ambient conditions, and a description of each experiment were stored in a native MATLAB file. Surface scans were also stored in a Wavefront OBJ file for later use.

\subsection{Data Processing and Analysis}
\label{subsec:dataProcessing}
\noindent
In order to improve data visualization, we applied a 50-point moving average filter to the recorded data. This filter is defined by the following equation whereby the average of every $l$ samples of the input data vector $x$ are computed and stored producing the output vector $y$.

\begin{equation}
y(n)=\frac{1}{l} \sum_{k=0}^{l-1} x(n-k)\;.
\end{equation}  

\begin{figure}[!hbt]
\centering
\includegraphics[width=0.6\linewidth]{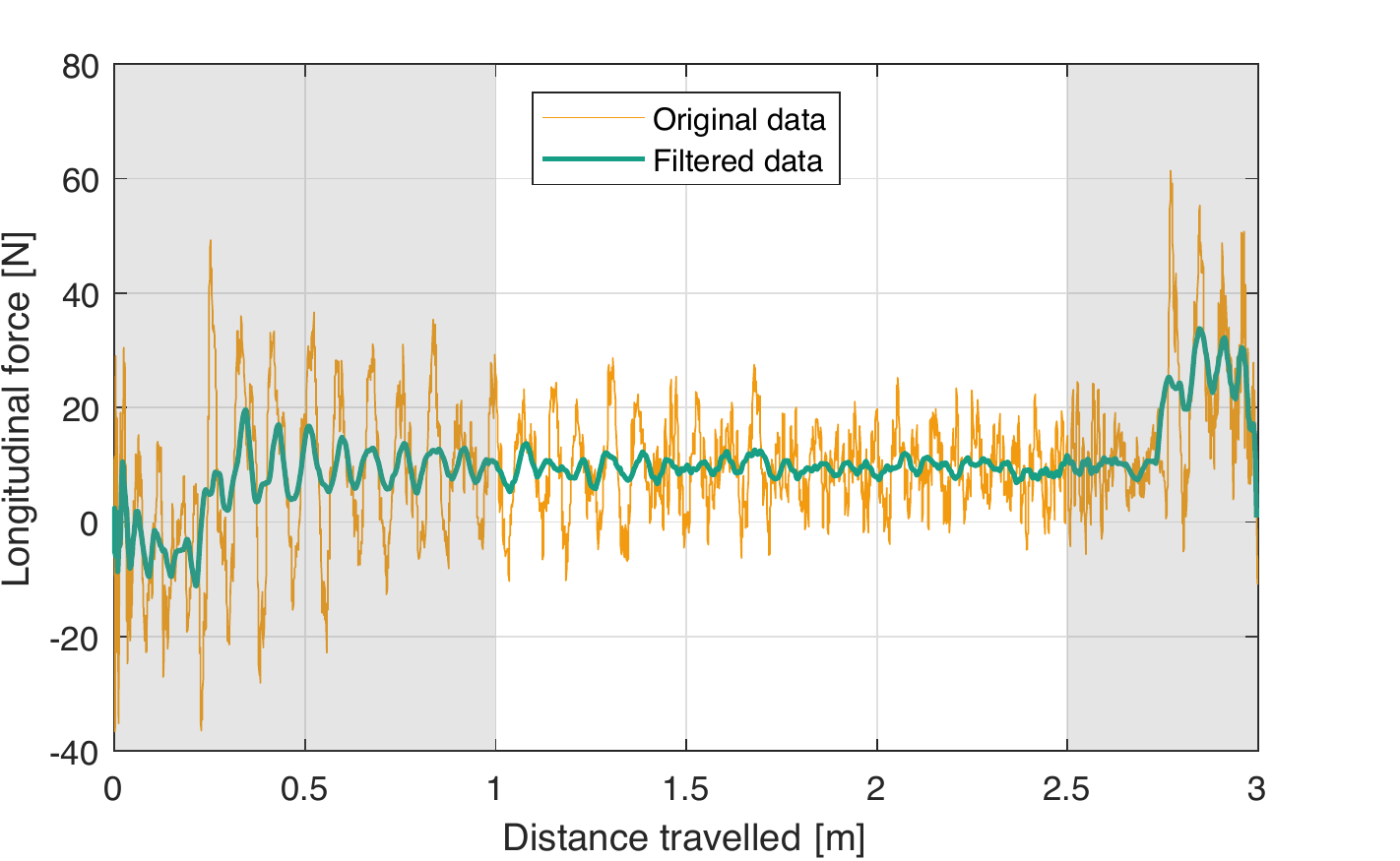}  
\caption{Original and filtered data. Data from transient states (gray-shaded regions) were removed from the analysis.}
\label{fig:filterData}
\end{figure}

The number of samples averaged is referred to as the window length of the filter. These type of linear filters are frequently applied to reduce the level of noise in time-varying input data. An example of the effect that the application of this filter has to the visualization of data can be seen in Fig.~\ref{fig:filterData}. Here, the variation of the longitudinal force the wheel is subjected to as measured by the force-torque sensor is drawn along the driving distance. Both original and filtered data are displayed.

In addition to the initial filtering of the input signals, measurements from transient states (initial acceleration, stabilization of vertical load, and deceleration) are removed prior to compute the average values of each individual experiment. A sensitivity analysis was conducted to set the limits along the driving distance between which the minimum standard deviation was reached. These limits, which are also depicted in Fig.~\ref{fig:filterData}, were found to be located at 1 and 2.25\,m (33\% and 75\% of the driving distance, respectively) for all the cases under study. Lastly, it should be noted that wheel sinkage was computed relative to the vertical location of the robot flange when an initial contact force of 10\,N was recorded during the loading process. 

\section{Results}
\label{sec:results}
\noindent
In the following sections we present experimental evidence of the velocity-dependent behavior of each of the case studies previously described. Two types of data analysis are conducted. Initially, the variation of key performance metrics with increasing speeds is evaluated for slip ratios of 20\% and 60\%. The selected performance metrics include: drawbar pull coefficient, wheel sinkage, and tractive efficiency. The drawbar pull coefficient, also refer by some authors as tractive ratio, defines the ratio of forward force (or drawbar pull) to normal load. Tractive efficiency, on the other hand, measures the percentage of the effective output power being ultimately used to drive a payload or an external load, represented in our case by the imposed wheel load. Tractive efficiency represents a more explicit indicative of the tractive performance of a vehicle \citep{Creager2017}. Efficiency is calculated based on the following equation \citep{Meyer1977}.

\begin{equation}
\eta = \frac{P_{db}}{P_{e}} = \frac{F_x v_x}{T_m \omega}\;,
\end{equation}

where $ P_{db}$ is the drawbar power, defined by the product of forward force (drawbar) as measured by the force-torque sensor, $F_x$, and the longitudinal velocity of the wheel, $v_x,$; and $P_e$ is the effective power output provided from the drive unit, defined by the product of output torque of the motor, $T_m$, and its rotational velocity, $\omega$.

In a second analysis, the evolution of drawbar pull and tractive efficiency are evaluated for each individual speed tested along a wider range of slip conditions.

In all of the following figures, data points represent averaged values of each individual experiment taking into consideration the processes described in Sect.~\ref{subsec:dataProcessing}. Error bars represent one standard deviation of the original, unfiltered data. Alongside the measurements gathered during the experiments, curves are displayed, which were estimated using a biharmonic spline interpolation method. 

\subsection{Case Study 1: Rigid Grousered Wheel}
\begin{figure}[!hbt]
\subfloat[\label{subfig1:class1A}]{%
\includegraphics[width=0.32\linewidth]{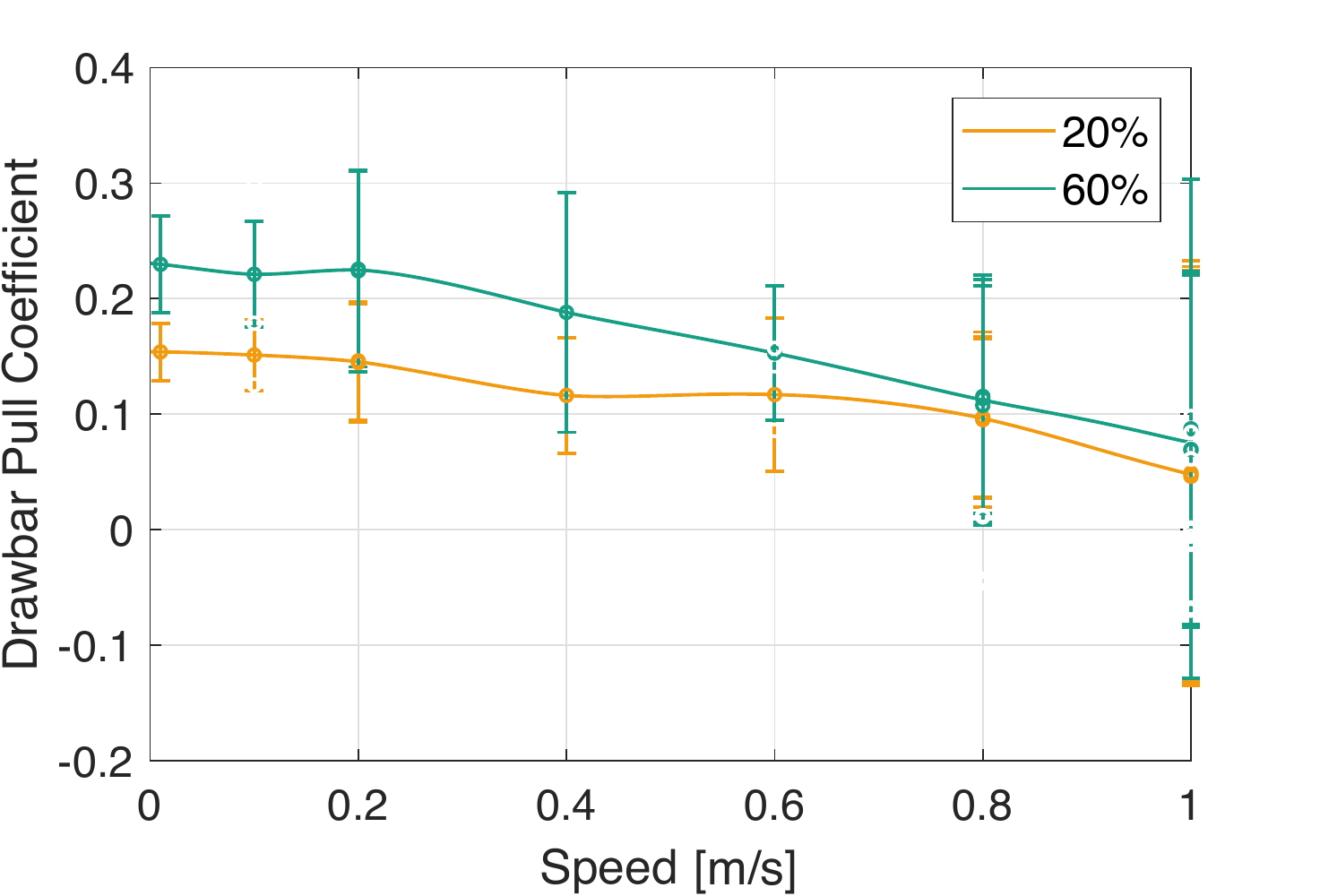}}
~
\subfloat[\label{subfig2:class1A}]{%
\includegraphics[width=0.32\linewidth]{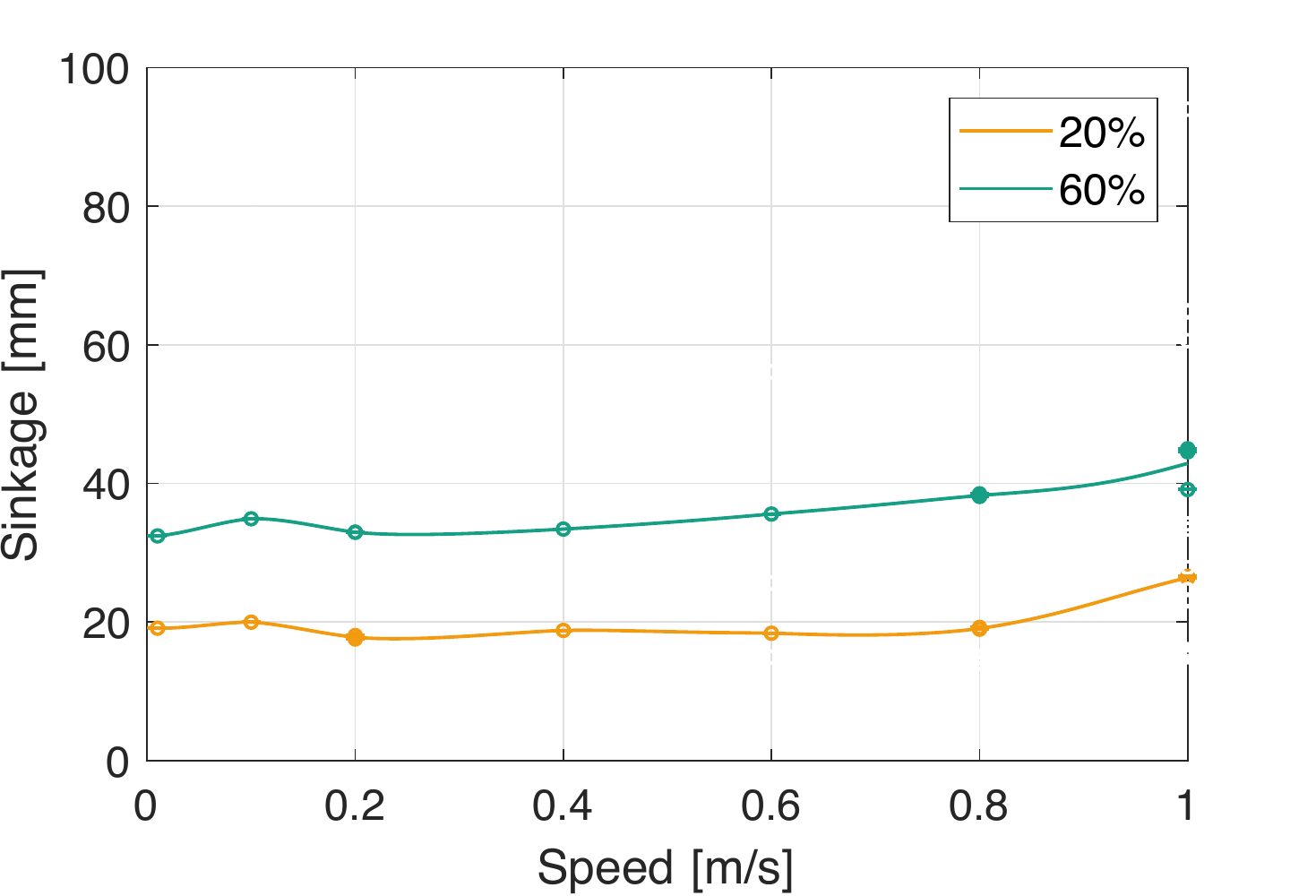}}
~
\subfloat[\label{subfig3:class1A}]{%
\includegraphics[width=0.32\linewidth]{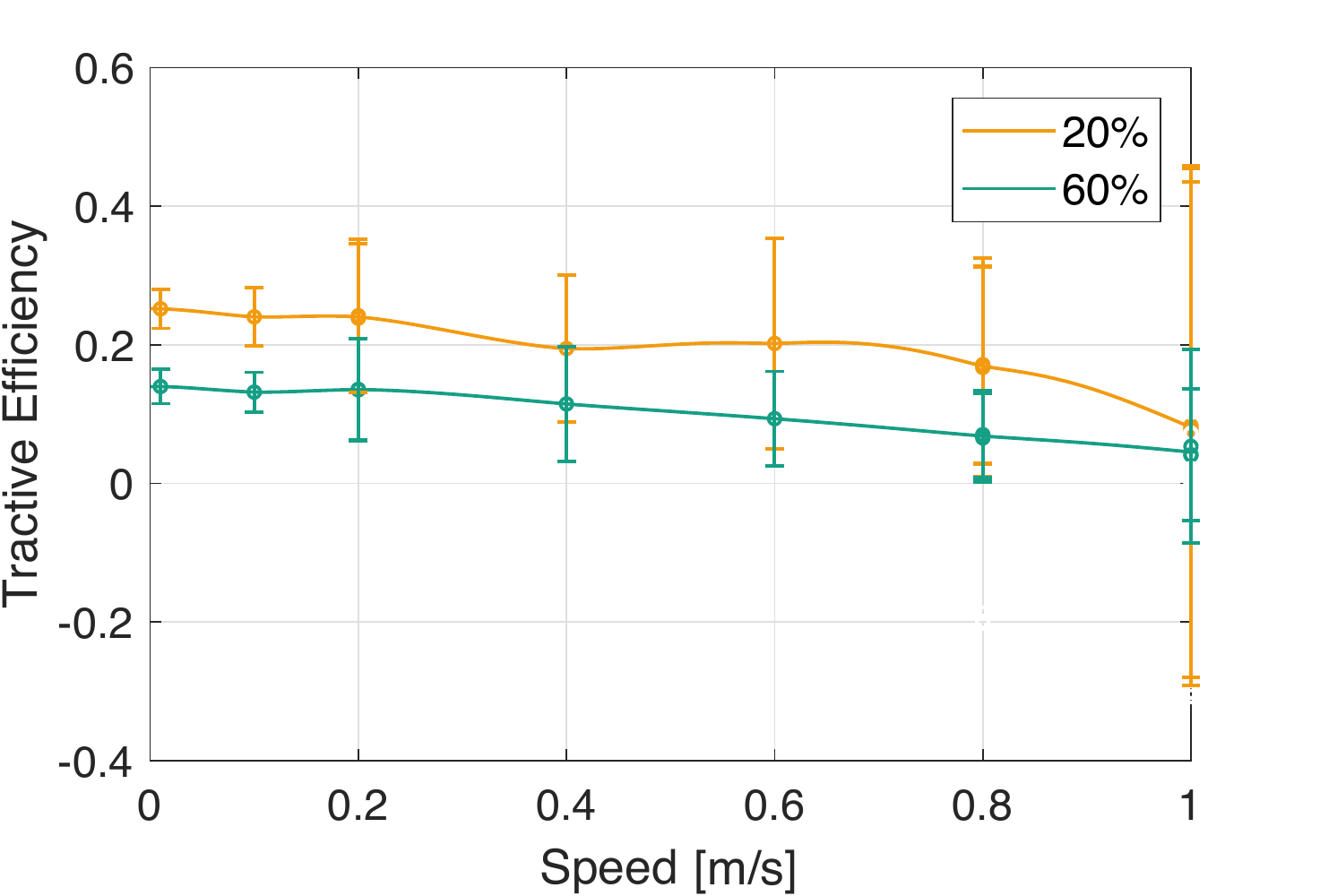}}
\caption{Measurements of the variation of (a) drawbar pull coefficient, (b) wheel sinkage, and (c) tractive efficiency with speed at slip ratios 20\% and 60\% for the case of a rigid grousered wheel of 250\,mm in diameter, 112\,mm wide, and 12 toothed grousers, subjected to a load of 100\,N, operating over olivine-based sand.}
\label{fig:class1A}
\end{figure}

\noindent
The variation with speed of the selected performance metrics for slip ratios of 20\% and 60\% are presented in Fig.~\ref{fig:class1A}. Although hardly visible in some cases due to overlap, measurements from multiple experiment conducted at the same speed--slip conditions are displayed. Overall, the drawbar pull coefficient decreased with increasing velocities for both low and high slip conditions, this reduction being more pronounced for speeds higher than 0.2\,m/s. At 20\% slip, the loss of drawbar pull was only 8\% for speeds between 0.01 and 0.2\,m/s, whereas this reduction was 63.7\% for speeds between 0.2 and 1\,m/s (Fig.~\ref{fig:class1A}(a)). Although lower drawbar pull values were measured at higher velocities, the higher the speed, the less influence the slip ratio had on the reduction of drawbar pull. At 0.2\,m/s, the reduction of the drawbar pull from high to low slip conditions was 2.5 times greater than that experienced at 0.8\,m/s. A higher variation in the averaged values was found at higher velocities and higher slip ratios; although these variations only seemed appreciable at 1\,m/s. The largest variability in the data was also observed a 1\,m/s (with drawbar pull coefficient values of $0.0523\pm0.1803$ at 20\% slip and $0.0802\pm 0.1738$ at 60\% slip). 

The wheel sinkage increased with increasing speeds but it was kept almost constant at 20\% slip for speeds between 0.2 and 0.8\,m/s (Fig.~\ref{fig:class1A}(b)). A reduction in sinkage of 2 mm was experienced when the speed was increased from 0.1 to 0.2\,m/s for both 20\% and 60\% slip conditions. As with drawbar pull measurements, minimum variability was attained in the multiple recorded trajectories at equal velocities. 

Tractive efficiency decreased with speed, but almost constant values were measured between 0.1 and 0.2\,m/s (Fig.~\ref{fig:class1A}(c)). A similar reduction in efficiency of approximately 70\% was experienced along the full range of velocities under both low and high slip conditions.

\begin{figure}[!hbt]
\subfloat[\label{subfig1:class1B}]{%
\includegraphics[width=0.535\linewidth]{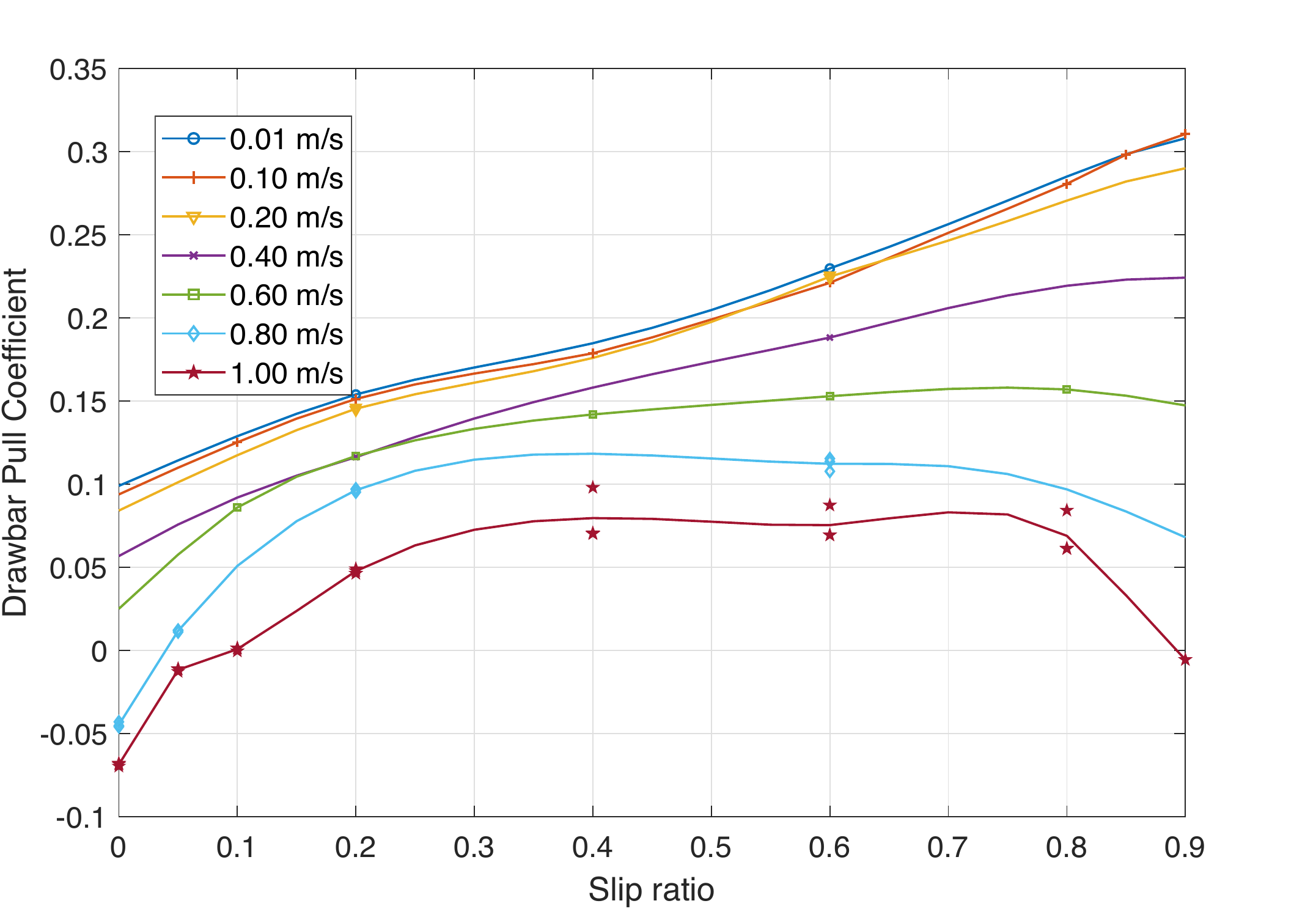}}
~
\subfloat[\label{subfig2:class1B}]{%
\includegraphics[width=0.465\linewidth]{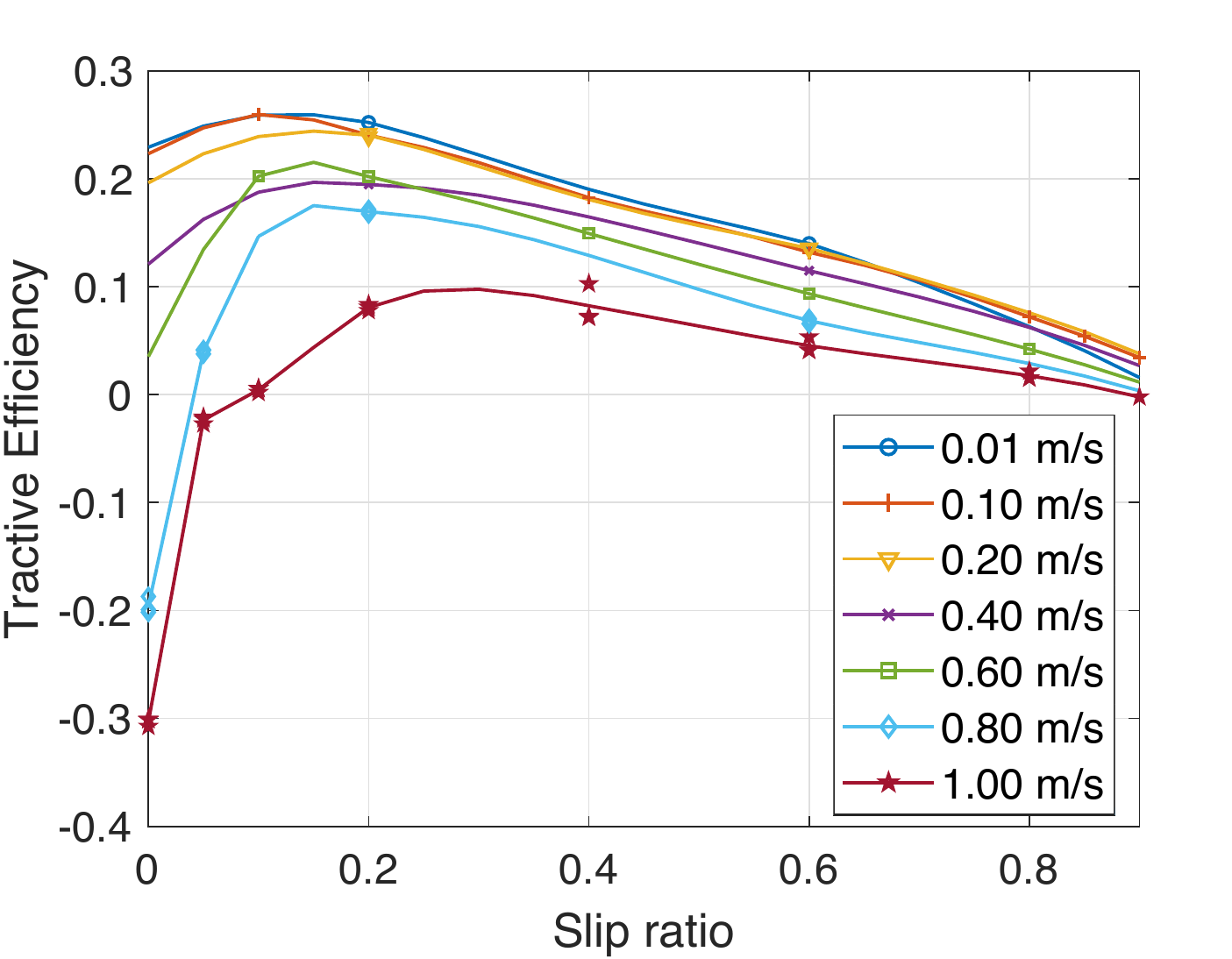}}
\caption{Measurements of the variation of (a) drawbar pull coefficient and (b) tractive efficiency with slip at each individual speed tested for the case of a rigid grousered wheel of 250 mm in diameter, 112 mm wide, and 12 toothed grousers, subjected to a load of 100 N, operating over olivine-based sand.}
\label{fig:class1B}
\end{figure}

Measurements of the variation of drawbar pull and tractive efficiency over the full range of slip ratios are presented in Fig.~\ref{fig:class1B}. The drawbar pull coefficient followed a nearly-constant increasing rate at speeds below 0.4\,m/s (Fig.~\ref{fig:class1B}(a)). A tendency to plateau was observed for speeds over 0.6\,m/s at slip ratios greater than 20\%. Test runs conducted at 1\,m/s and 90\% slip experienced a drastic reduction in drawbar pull, as opposed to the constant increment observed at 0.1\,m/s under the same slip conditions. With regards to the evaluation of tractive efficiency, similarly to the observations made with the variation of drawbar pull in the previous analysis, the higher the slip ratio, the less influence speed had on the reduction of tractive efficiency (Fig.~\ref{fig:class1B}(b)). Interestingly, the slip ratio at which maximum efficiency was reached appeared to increase with increasing speeds. At 0.1\,m/s maximum tractive efficiency was achieved at 10\% slip, whereas maximum efficiency at 1\,m/s was achieved at 30\% slip. Maximum tractive efficiency decreased 62.5\% when speeds were increased from 0.01 to 1\,m/s. Efficiency decreased almost linearly after the maximum was reached for every speed tested.

\subsection{Case Study 2: Flexible Grousered Wheel}
\begin{figure}[!hbt]
\subfloat[\label{subfig1:class2A}]{%
\includegraphics[width=0.32\linewidth]{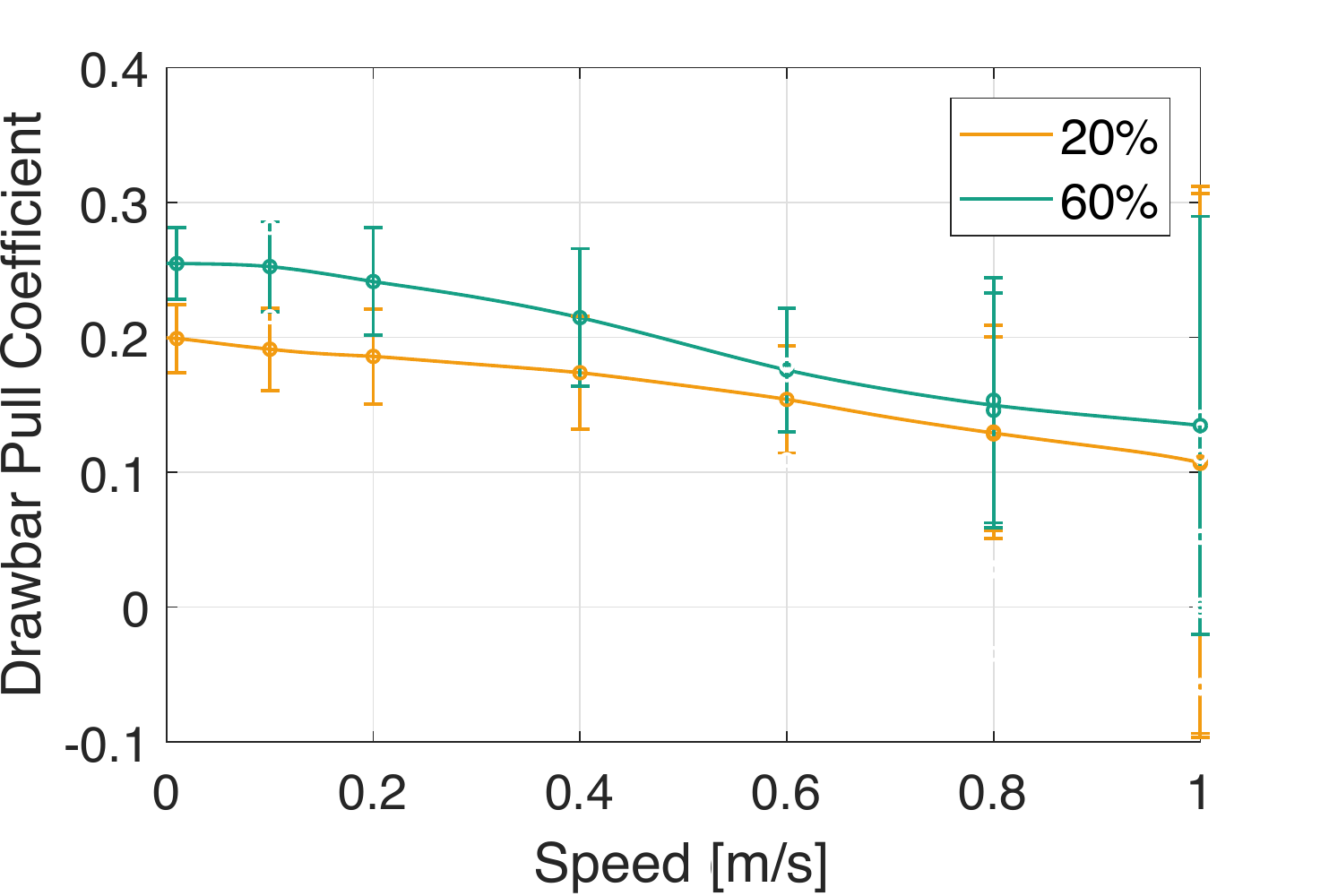}}
~
\subfloat[\label{subfig2:class2A}]{%
\includegraphics[width=0.32\linewidth]{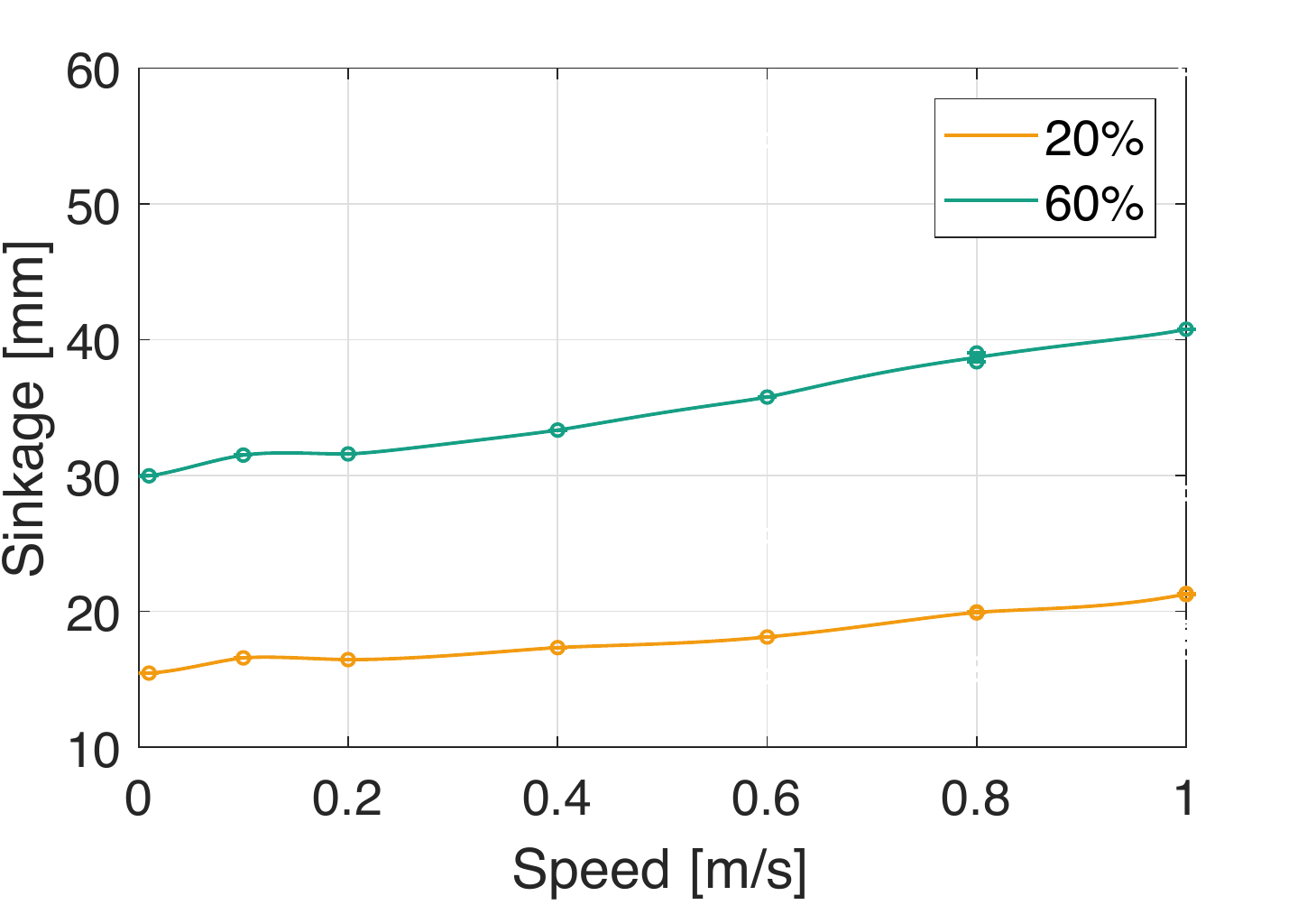}}
~
\subfloat[\label{subfig3:class2A}]{%
\includegraphics[width=0.32\linewidth]{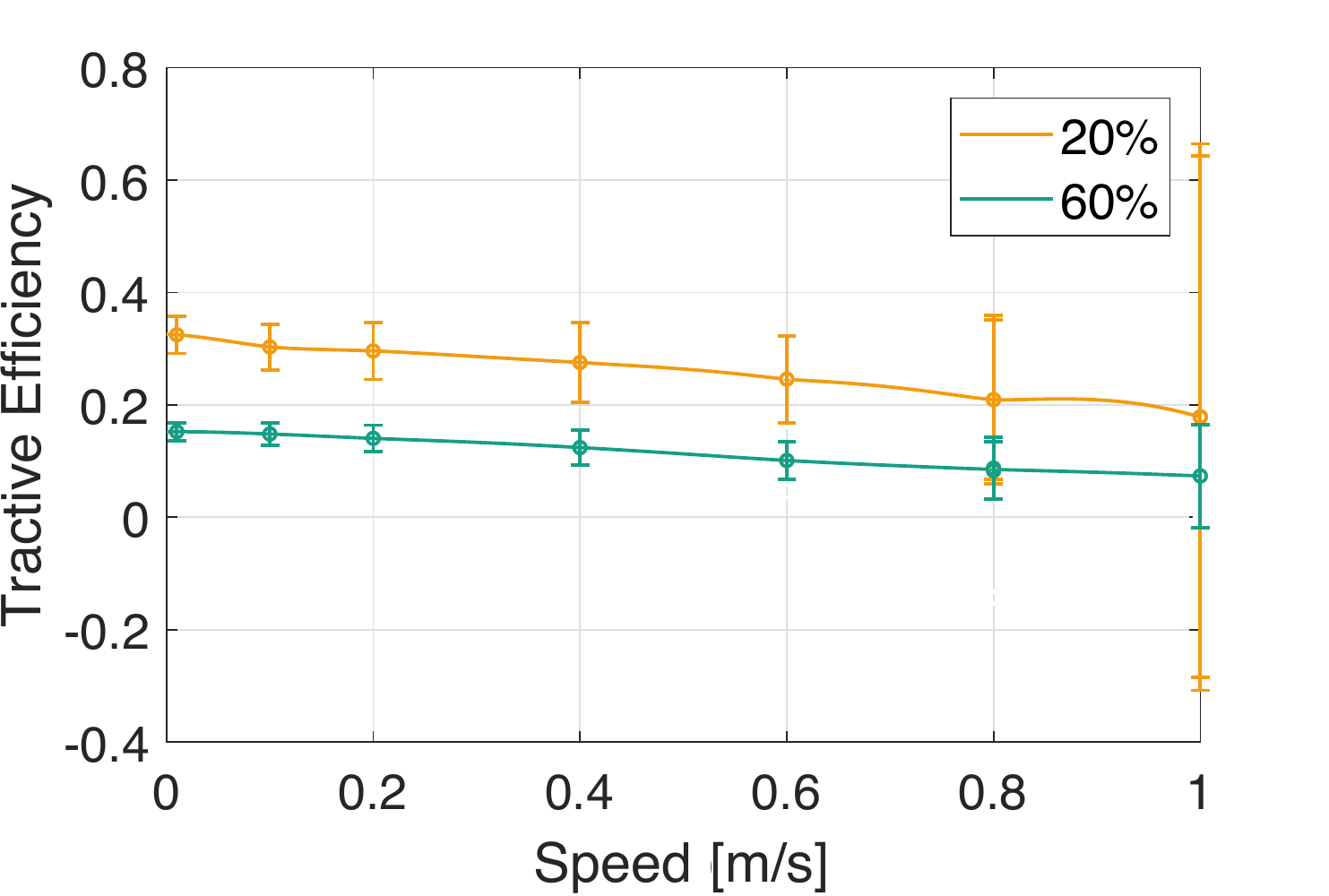}}
\caption{Measurements of the variation of (a) drawbar pull coefficient, (b) wheel sinkage, and (c) tractive efficiency with speed at slip ratios 20\% and 60\% for the case of a flexible metallic wheel of 250\,mm in diameter, 112\,mm wide, and 12 toothed grousers, subjected to a load of 100\,N, operating over olivine-based sand.}
\label{fig:class2A}
\end{figure}

\noindent
Very similar trends were observed in this second case study. Drawbar pull and tractive efficiency decreased while sinkage increased with increasing speeds (Fig.~\ref{fig:class2A}). When comparing relative variations between high and low slip conditions and high and low speeds, data display a very similar response to that observed in the previous case. The rate at which sinkage increased with speed doubled at 60\% slip (Fig.~\ref{fig:class2A}(b)). As with our first case, a minimum reduction in sinkage was experienced between 0.1 and 0.2\,m/s. It should be noted that in the case of the flexible wheel, sinkage was computed taking into account a measured maximum wheel deflection of 5\,mm under load. In this second case, efficiency decreased approximately 45\% for speeds varying from 0.01 to 1\,m/s in both low and high slip conditions (Fig.~\ref{fig:class2A}(c)).

\begin{figure}[!hbt]
\subfloat[\label{subfig1:class2B}]{%
\includegraphics[width=0.5\linewidth]{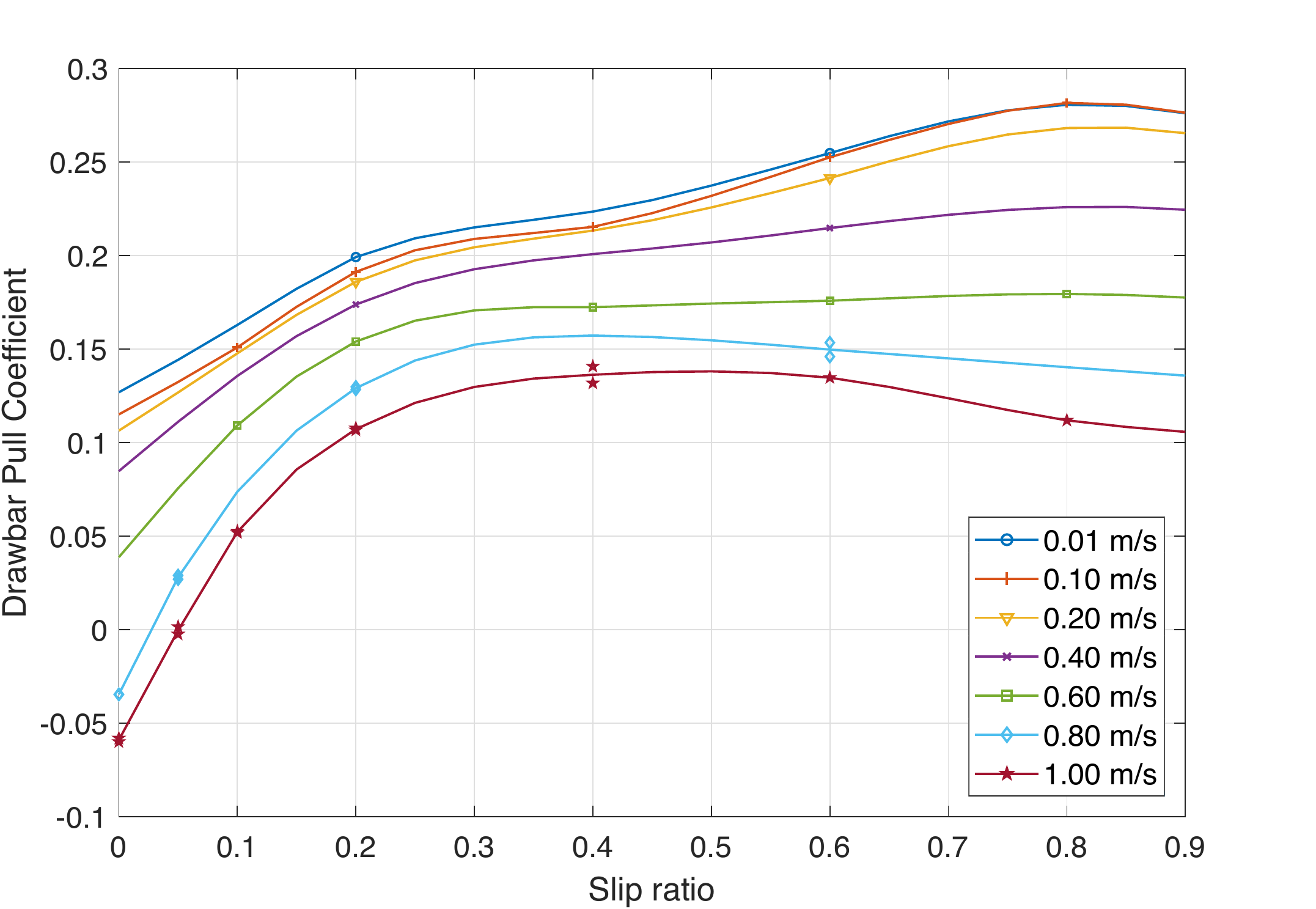}}
~
\subfloat[\label{subfig2:class2B}]{%
\includegraphics[width=0.5\linewidth]{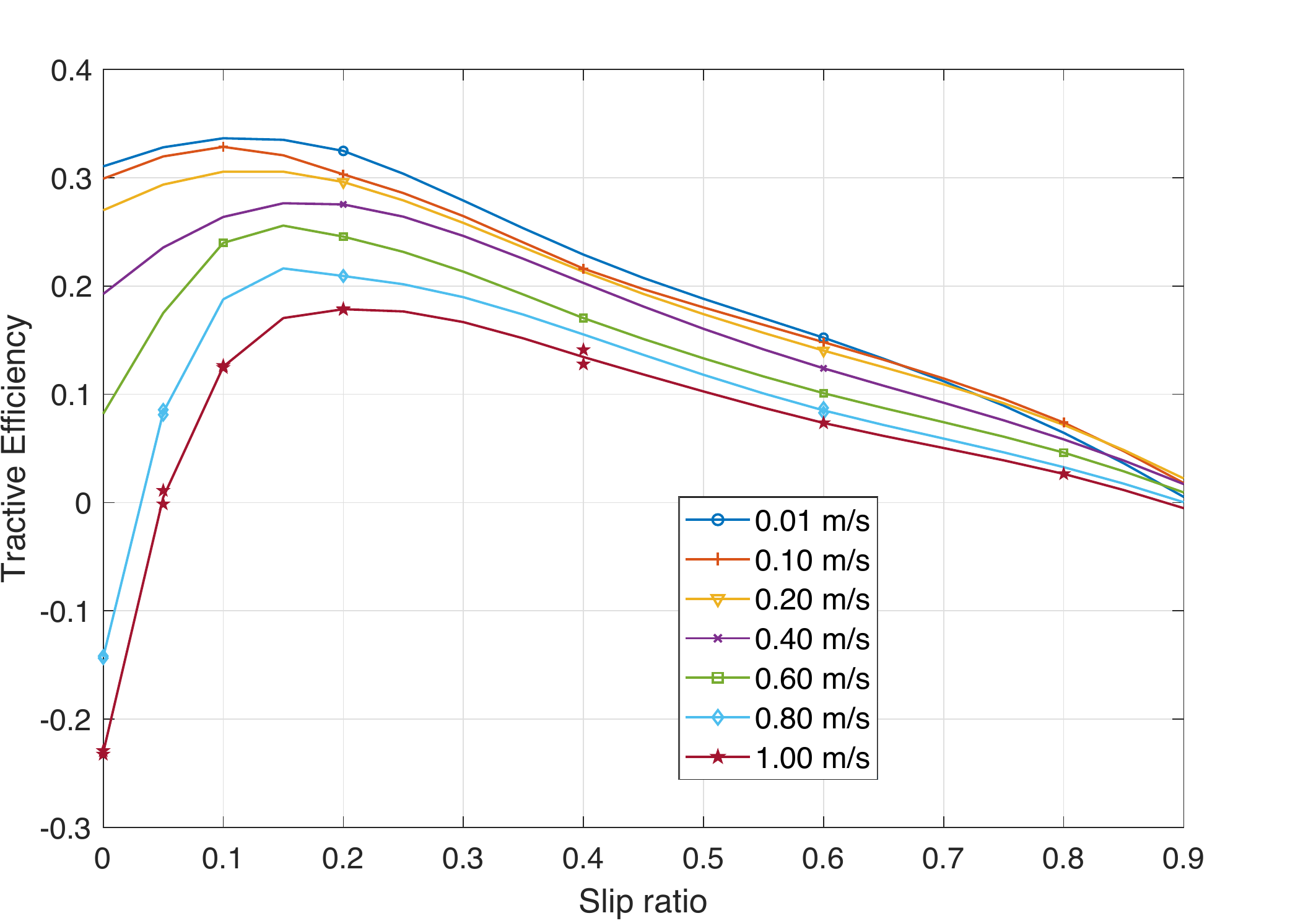}}
\caption{Measurements of the variation of (a) drawbar pull coefficient and (b) tractive efficiency with slip at each individual speed tested for the case of a flexible metallic wheel of 250 mm in diameter, 112 mm wide, and 12 toothed grousers, subjected to a load of 100 N, operating over olivine-based sand.}
\label{fig:class2B}
\end{figure}

The variation of drawbar pull with slip display a similar tendency to plateau as speeds approach 0.6\,m/s (Fig.~\ref{fig:class2B}(a)). Measurements at 90\% slip could not be conducted in this case due to an excessive sinkage of the wheel. The attachment points of the flexible wheel to the drive unit differed from that of the previous case. The design of the center hub of the flexible wheel made it easier for it to get in contact with the soil surface when higher sinkages were realized; thus, preventing its use under very high slip conditions. The slip ratio at which maximum efficiency was reached appeared to also increase with increasing velocities (Fig.~\ref{fig:class2B}(b)). At 0.2\,m/s, maximum efficiency was achieved at 10\% slip, whereas at 1\,m/s this was achieved at 20\%. A similar linear reduction in efficiency was observed after the maximum was reached for every speed tested.  

\subsection{Case Study 3: $\textrm{CaCO}_3$-based Soil}
\begin{figure}[!hbt]
\centering
\subfloat[\label{subfig1:class3A}]{%
\includegraphics[width=0.32\linewidth]{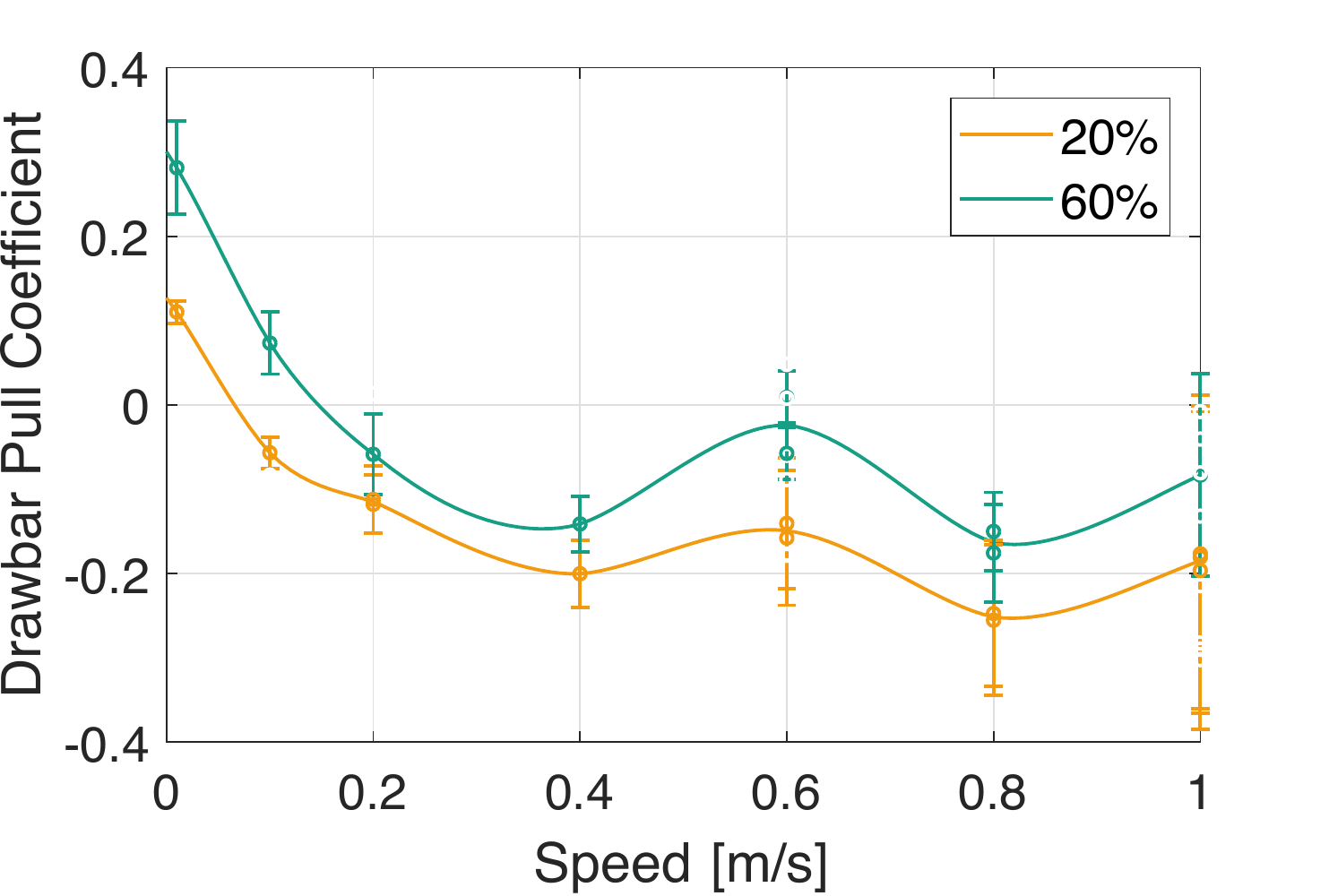}}
~
\subfloat[\label{subfig2:class3A}]{%
\includegraphics[width=0.32\linewidth]{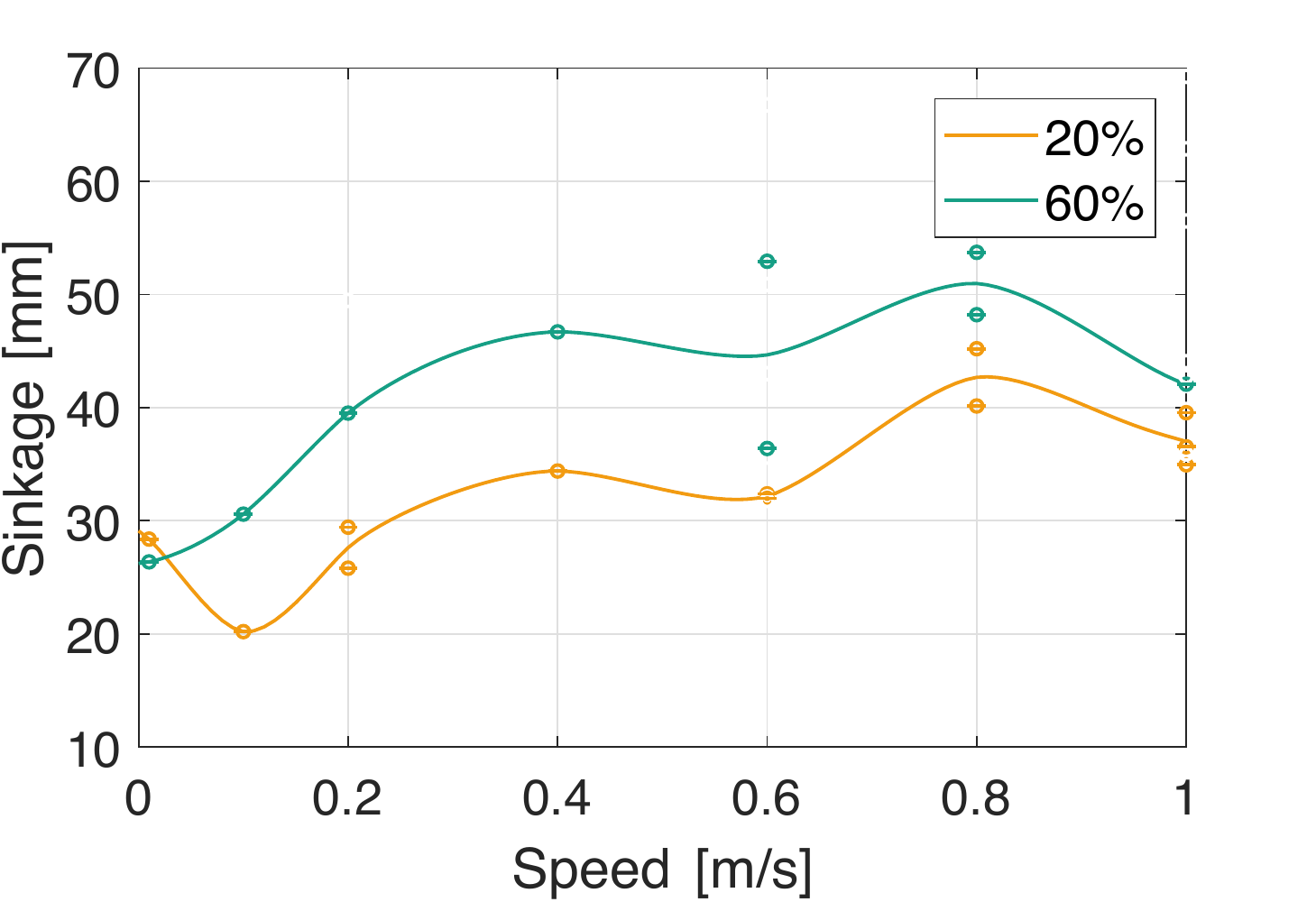}}
~ \\ \centering
\subfloat[\label{subfig3:class3A}]{%
\includegraphics[width=0.4\linewidth]{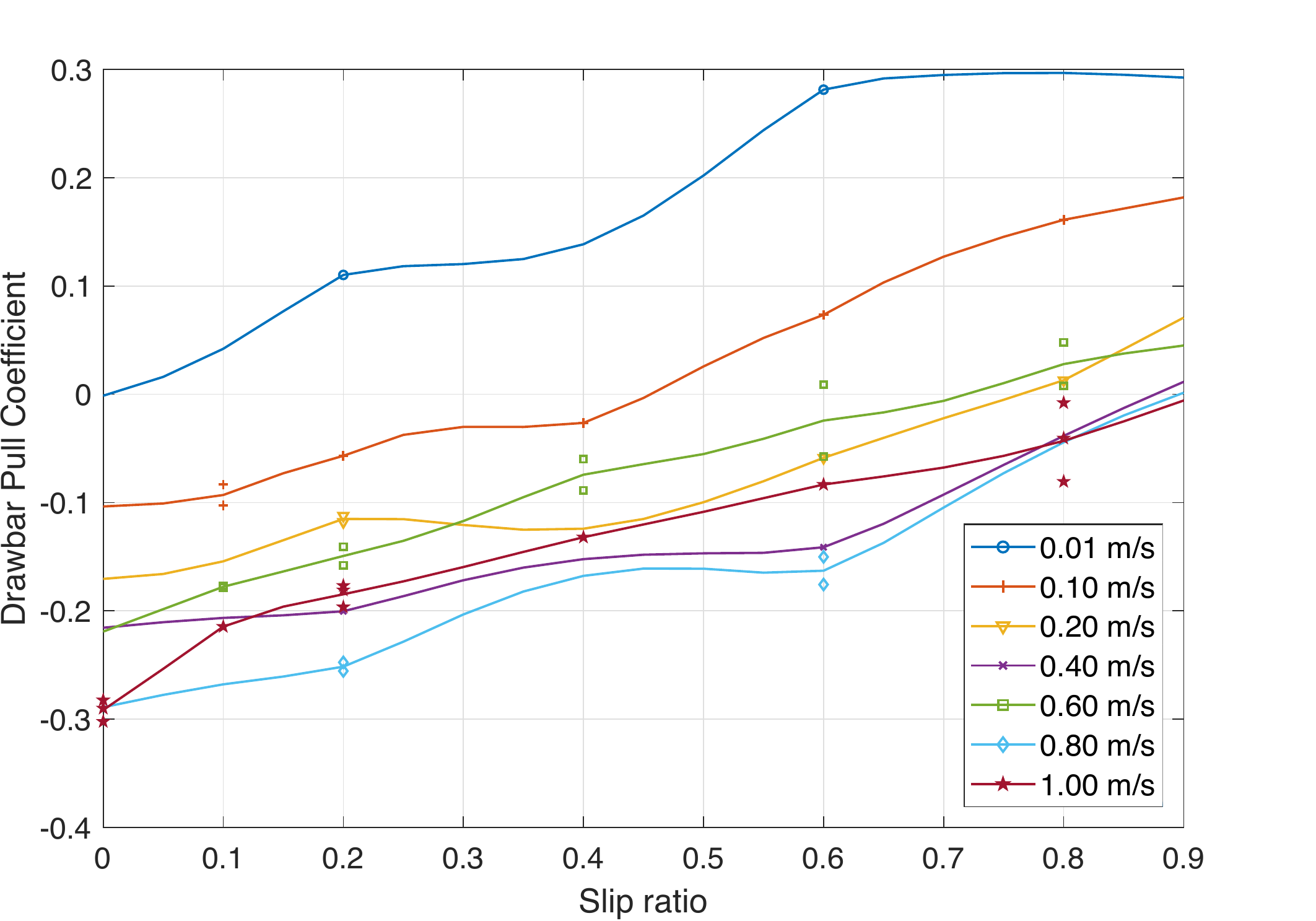}}
\caption{Measurements of the variation of (a) drawbar pull coefficient and (b) wheel sinkage with speed at slip ratios 20\% and 60\% ; and (c) variation of the drawbar pull coefficient with slip at each individual speed tested for the case of a rigid grousered wheel of 250\,mm in diameter, 112\,mm wide, and 12 toothed grousers, subjected to a load of 100\,N, operating over $\textrm{CaCO}_3$-based soil.}
\label{fig:class3A}
\end{figure}

\noindent
As already mentioned, the RMCS13 required a more laborious preparation method and higher degree of safety measures in its manipulation due to its slight cohesiveness and small grain size distribution. Unlike with the olivine-based sand, compression of the soil was necessary for the wheel to operate under load without excessively sinking on the ground. Even after taking these considerations into account, the operation of the wheel was found to be rather burdensome, unstable, and inconsistent for speeds that surpass 0.1\,m/s and slip ratios beyond 20\%. Despite the issues encountered, measurements gathered when the wheel was driven over the RMCS13 displayed no significant change in the overall trends (Fig.~\ref{fig:class3A}). Drawbar pull decreased and wheel sinkage increased with increasing velocities. The challenging operation of the wheel over RMCS13 can be inferred from the higher degree of variability and reduced stability of the collected measurements. A complete loss of traction was consistently experienced at speeds above 0.1\,m/s ---reflected in negative values of drawbar pull almost throughout the full range of speeds tested (Fig.~\ref{fig:class3A}(a)). The inability to generate enough traction was also experienced in close-to-zero input torques that precluded the computation of efficiency values. Sinkage no longer follows a constant increasing rate with speed, but appeared to decrease from 0.4--0.6\,m/s and from 0.8--1\,m/s (Fig.~\ref{fig:class3A}(b)). The variation of drawbar pull with slip is now presented in Fig.~\ref{fig:class3A}(c)). The challenging conditions imposed by the RMCS13 are also reflected here, where positive traction levels for the complete range of slip ratios were only achieved during the experiments conducted at 0.01\,m/s.

\section{Discussion}
\label{sec:discussion}
\noindent

A thorough characterization of the velocity-dependent performance of two commonly-used rover wheel designs over two different soils was conducted. The complete test campaign comprised 23 consecutive days of testing, in which more than 500 single-wheel experiments were performed and data were recorded over a total compounded distance of approximately 1.5\,km. Three key performance indicators throughout multiple case studies were monitored and part of the measurements collected have been presented above. 

As already addressed in the previous section, equal trends were observed among the different cases analyzed. The evaluation of tractive performance with varying slip conditions has been extensively revisited in the literature. Our measurements reflected equal relational behaviors. Higher drawbar pull values were measured at increased slip ratios \citep{Ishigami2006,Inotsume2014,Senatore2014,Ding2015} and a tendency to plateau at higher slip conditions was observed particularly at higher speeds \citep{Wong2008,Ding2011,Oikawa2017,Creager2017}. Absolute drawbar pull coefficient and tractive efficiency values were invariably measured below 0.3 and 35\%, respectively, for all the cases investigated. Tractive efficiency reached its maximum at low slip values, often between 10--30\%, and decreased almost linearly with increasing slip ratios for most of the velocities and cases evaluated. Similar outcomes have been previously presented in the literature \citep{Creager2017}. The slip ratio at which maximum efficiency was achieved was observed to increase with increasing speeds. 

With regards to the velocity-dependent relationships in which this study focuses, a lower drawbar pull and efficiency were measured at higher speeds. A lower tractive performance with increasing velocities was expected as was previously shown to be the case in the operation of farming trucks over unconsolidated soils \citep{Bauer2013,Polcar2017}. 

The flexible metallic wheel used in case study 2 displayed the best performance along the full range of velocities. Its greater contact surface area led to lower pressure values, lower sinkages, and a more efficient performance. The flexible wheel also displayed the lowest relative variation in the data between low and high speed ranges. Proof of a 10--30\% improvement in performance at low speeds when flexible wheels were used over unconsolidated soils can be found in the literature \citep{Favaedi2011}. Our experiments confirm and expand these observations to a larger speed range. Measurements gathered at 1\,m/s with the flexible wheel displayed over 2 times higher drawbar pull and tractive efficiency, and 18\% lower sinkage at 20\% slip compared to that of the rigid wheel.

Among the different speeds tested, the lowest impact on performance was experienced for speeds up to 0.2--0.3\,m/s. This implies speeds that are already 5--10 times higher than currently-conceived rover driving velocities. On average among the two wheel designs evaluated, between speeds of 0.01 and 0.2\,m/s, drawbar pull and tractive efficiency decreased 7.5\% and 7\%, respectively, and approximately equal sinkages were experienced. This is an important realization in the pursuit of decreasing rover transit times without negatively impacting performance. The impact that such an increase in rover driving velocity would have on a mission, in terms of risks and benefits from a general perspective, and on the design of the locomotion subsystem and how it would affect current control and navigational approaches, in particular, are still to be evaluated.  

\begin{figure}[!hbt] 
\centering
\includegraphics[width=0.6\textwidth]{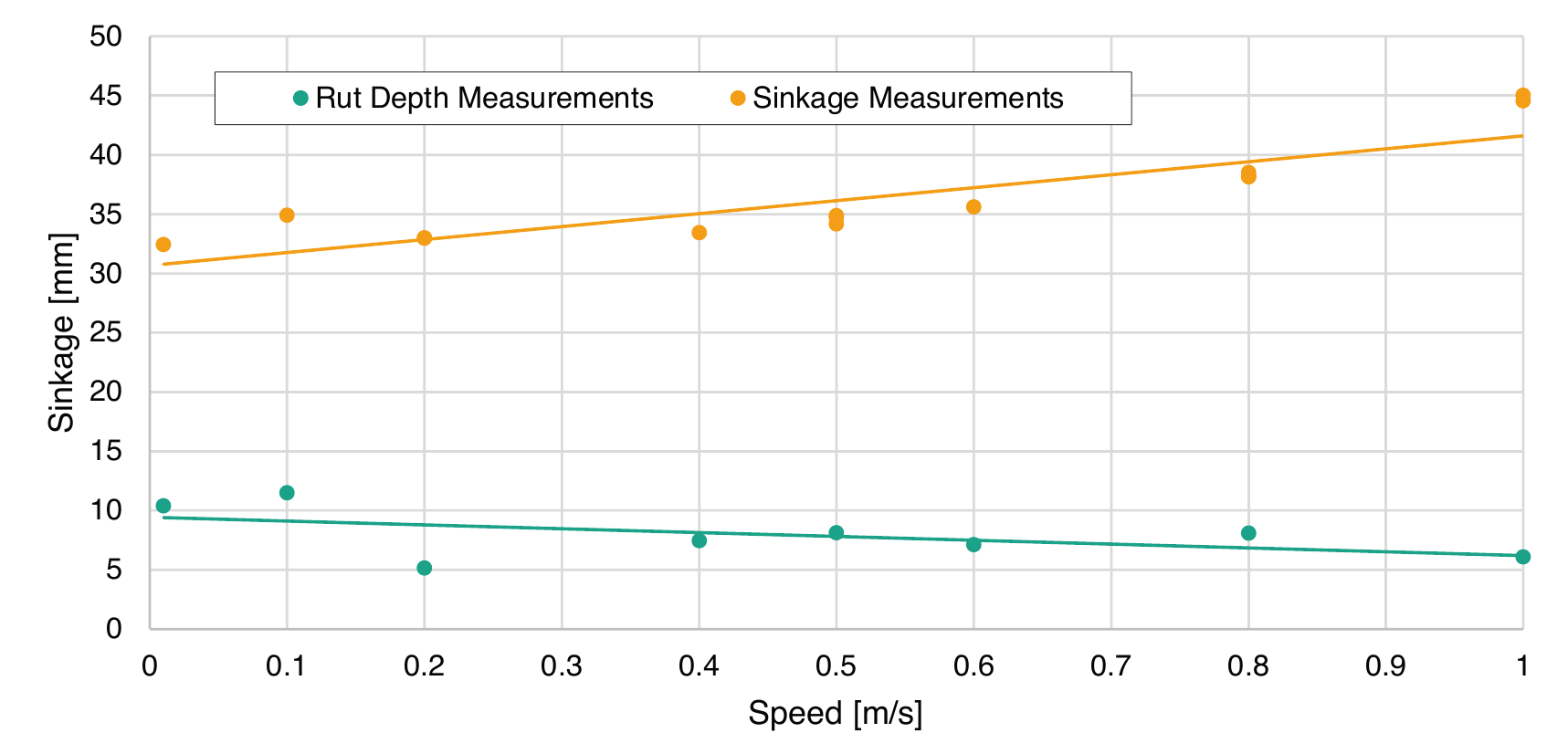}
\caption{Comparison between ruth depth and sinkage measurements at 1\,m/s and 60\% slip for a rigid grousered wheel.}
\label{fig:rutdepth}
\end{figure}

Unlike what previous experimental evidence suggested, no significant change in data trends with speed was observed for any of the metrics evaluated in our experiments. Overall, wheel sinkage increased with increasing speeds showing no clear indication of the appearance of soil lift. Multiple reasons may explain the latter. First and foremost, the phenomenon of soil lift was first suggested during the operation of airplanes over unpaved runways \citep{Crenshaw1972}. In this context, wheel loads 100 times heavier and velocities 30--40 times higher were being considered. The conditions limiting our experiments (i.e., maximum driving distance and top speed) might have been excessively restrictive for the so-called planing velocity to be triggered ---i.e., the minimum speed required for soil lift to occur. Additionally, partial evidence for the appearance of soil lift have been presented through experiments that were conducted over extremely shallow depths of soil, some times even just as low as a few centimeters \citep{Coutermarsh2007}. Given the heavy wheel loads enforced in these studies, it could be argued that relevant dynamic distortions were taking place due to the proximity of solid layers of asphalt and concrete. Lastly, in most of these studies the manifestation of soil lift effects had been solely justified based on a postmortem analysis of the trail marks left by the wheels \citep{Crenshaw1972,Beaty1975,Coutermarsh2007}. As was observed during our experiments, soil transportation effects were becoming increasingly predominant as velocity and slip ratio increased. Larger amounts of soil were displaced and sprayed onto the wheel track by effect of the grousers and the increased rotational velocity of the wheel. This, together with the slight elastic recovery of the soil, makes rut depth measurements alone a potential source for misinterpreting overall effects on wheel sinkage. Figure~\ref{fig:rutdepth} shows the data trends followed by the evolution of wheel sinkage, as registered by the vertical position of the robot flange, and rut depth measurements obtained from the recorded surface scans. Data trends followed opposite directions when high speeds and high slip ratios were considered.

Another potential reason that could explain the absence of turning points in the evolution of performance metrics with speed lies on technical constraints concerning our experimental setup. Most of the experiments cited in the literature were conducted on free-rolling, towed wheels, whereas in our case the while was driven by its own drive unit and a given slip ratio was imposed by a robotic arm. The main controller of the robot arm had to constantly compensate for the rotational speed of the wheel, which proved to be increasingly difficult when higher speeds or higher slip ratios were considered. A higher variability and data dispersion under these conditions were observed. Greater sinkages may have been experienced due to slightly over-imposed slipping conditions resulting from the controller throughput being challenged to keep pace with the motor driver. Further limitations and potential sources of disturbance in our approach include the already-mentioned 3-m limited driving distance, the close approach in several instances to the motor current limit (particularly at high speeds), and to slightly higher vibrations experienced at both ends of the soil bin where the robot arm was fully extended. Whether the aforementioned beneficial effects of soil lift could be triggered within the range of loads and speeds considered in planetary exploration is still unknown. 

\begin{figure}[!hbt] 
\centering
\includegraphics[width=\linewidth]{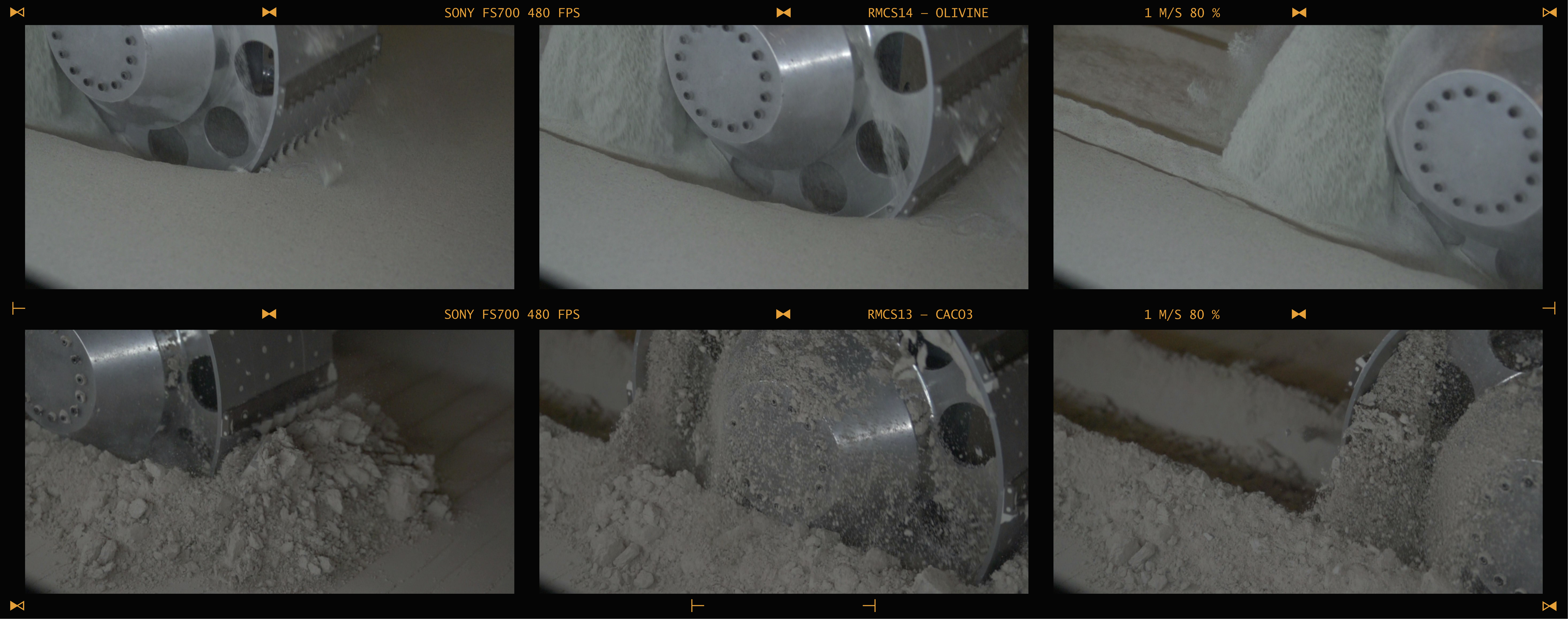}
\caption{A rigid grousered wheel operating over two different soils: an olivine-based sand called RMCS14 (above) and a $\textrm{CaCO}_3$-based soil called RMCS13 (below).}
\label{fig:480}
\end{figure}

Finally, a few observations are worth highlighting regarding the performance over different types of soils. Besides the aspects already mentioned regarding the more laborious manipulation of and rather challenging operation over the $\textrm{CaCO}_3$-based soil, the observable soil behavior varied greatly for the two soils evaluated. Still frames from the footage recorded of the interaction of our baseline wheel with RMCS14 and RMCS13 soils at 1\,m/s and 80\% slip are displayed in Fig.~\ref{fig:480}. Differences in the wheel-soil interactive behavior are clearly visible. Above a speed of 0.4\,m/s, the macroscopic behavior of the olivine-based sand (RMCS14) started to display an increased fluidity that resembled that of a liquid. This flowing capacity was unnoticed in the case of the $\textrm{CaCO}_3$-based soil (RMCS13), where an increased accretion of soil on the front and sides of the wheel was observed. 

A careful analysis of the behavior of different soils and their interactive behavior with fast rotating wheels is still required. However, our experiments suggest that wheel design could be ultimately influenced by this transition in the macroscopic behavior of soils if an optimized performance is to be achieved at increasing velocities or high slip conditions. Frequently-encountered critical issues when operating over soft, unconsolidated terrains include a sudden loss of traction that may derive, in the worst case, in the entrapment of the wheels. This situation could be potentially avoided by wheel designs that favor dynamic lift generation over buoyancy under these conditions. Additionally, efficiency could be increased by the use of closed wheel designs (use of sidewalls or shallow dish wheels) that prevent the incessant entry of sand or favors its flow, which would also prevent small rocks and pebbles from impacting and damaging internal parts of the wheel. 

This study has provided a first glance into the potential effects that increasing velocity has on the tractive performance of planetary rovers. The data collected will serve in the evaluation of future steps in the development of fast-moving locomotion subsystems. At the same time, the results presented here constitute a map of the evolution and relationships among key tractive performance metrics, which could be used to formulate new and improved dynamic traction models. Other aspects associated with locomotion performance such as suspension dynamics also need to be evaluated either through multi-body simulations or field testing, both of which could also be greatly assisted by the results provided in this study.

\subsubsection*{Acknowledgement}
\noindent
This research was partially supported by the Japanese Ministry of Education, Culture, Sports, Science and Technology (MEXT) and Tohoku University's GP-Mech Program for the promotion of international collaborations in academia. We would like to express our gratitude to Dr. Johann Bals, Dr. Roy Lichtenheldt, and Rainer Krenn from the Institute of System Dynamics and Control of the German Aerospace Center (DLR) for their generosity and hospitality, which made this collaboration possible. We also wish to extend our appreciation to our colleagues from the Automation and Robotics Section of the European Space Research and Technology Centre (ESTEC/ESA). Special thanks goes to Gianfranco Visentin and Martin Azkarate, who provided insight and expertise that greatly assisted this research. 
\bibliographystyle{elsarticle-harv} 
\bibliography{new_bib}

\end{document}